\documentclass[11pt]{article}

\usepackage[utf8]{inputenc}
\usepackage{geometry}
\usepackage{subcaption}
\usepackage{amsmath, amssymb, amsfonts}
\usepackage{graphicx}
\usepackage{xurl}
\usepackage[colorlinks=true,
            linkcolor=black,
            urlcolor=black,
            citecolor=black,
            breaklinks=true,
            filecolor=black]{hyperref}
\usepackage{booktabs} 
\usepackage{xcolor}   
\usepackage{titlesec} 
\usepackage{enumitem} 
\usepackage{cite}
\usepackage[table]{xcolor}
\usepackage{cleveref}
\usepackage{wrapfig}
\usepackage{booktabs}
\usepackage{titletoc}
\usepackage{makecell}
\usepackage{rotating}
\usepackage{titling}
\usepackage{authblk}
\usepackage{booktabs, multirow} 
\usepackage{hyperref}
\usepackage[T1]{fontenc}

\crefname{section}{Section}{Sections}
\Crefname{section}{Section}{Sections}
\crefname{figure}{Figure}{Figures}
\Crefname{figure}{Figure}{Figures}
\crefname{table}{Table}{Tables}
\Crefname{table}{Table}{Tables}
\crefname{equation}{Equation}{Equations}
\Crefname{equation}{Equation}{Equations}

\newcommand\nnfootnote[1]{%
  \begin{NoHyper}
  \renewcommand\thefootnote{}\footnote{#1}%
  \addtocounter{footnote}{-1}%
  \end{NoHyper}
}

\makeatletter
\renewcommand\AB@affilsepx{, \protect\Affilfont}
\makeatother

\renewcommand\Affilfont{\footnotesize}

\title{\textbf{MIMIC: A Generative Multimodal\\Foundation Model for Biomolecules}}

\author[1,2]{Siavash Golkar$^\dagger$}
\author[1,3]{Jake Kovalic$^{*\dagger}$}
\author[1,2]{Irina Espejo Morales$^{*\dagger}$}
\author[4,5,9]{Samuel Sledzieski$^{*\dagger}$}
\author[4,5]{Minhuan Li$^{*\dagger}$}
\author[5,9]{Ksenia Sokolova$^\dagger$}
\author[1,7]{G\'eraud Krawezik$^\dagger$}
\author[1,4]{Alberto Bietti$^\dagger$}
\author[2]{Claudia Skok Gibbs}
\author[16]{Roman Klypa}
\author[17]{Shengwei Xiong}
\author[17]{Amy Crawford}
\author[1,10]{Fran\c{c}ois Lanusse}
\author[1,11]{Liam Parker}
\author[1,2]{Kyunghyun Cho}
\author[1,12,13]{Miles Cranmer}
\author[1,13]{Tom Hehir}
\author[1,2]{Michael McCabe}
\author[1]{Lucas Meyer}
\author[1,4]{Rudy Morel}
\author[1,12]{Payel Mukhopadhyay}
\author[1,15]{Mariel Pettee}
\author[1,6]{Helen Qu}
\author[1,18]{Jeff Shen}
\author[1,2]{David Fouhey}
\author[13,14]{Hadi Sotoudeh}
\author[5]{Vikram Mulligan}
\author[4,5]{Pilar Cossio}
\author[4,5]{Sonya M. Hanson}
\author[17]{Alisha N. Jones}
\author[5,8,9,19]{Olga G. Troyanskaya}
\author[1,2,6,18,20]{Shirley Ho\vspace{5pt}}

\affil[1]{Polymathic AI}
\affil[2]{Center for Data Science, New York University}
\affil[3]{Department of Applied Physics, Yale University}
\affil[4]{Center for Computational Mathematics, Flatiron Institute}
\affil[5]{Center for Computational Biology, Flatiron Institute}
\affil[6]{Center for Computational Astrophysics, Flatiron Institute}
\affil[7]{Scientific Computing Core, Flatiron Institute}
\affil[8]{Department of Computer Science, Princeton University}
\affil[9]{Princeton Precision Health, Princeton University}
\affil[10]{AIM, Universit\'e Paris-Saclay, Universit\'e Paris Cit\'e}
\affil[11]{Department of Physics, University of California}
\affil[12]{Department of Applied Mathematics and Theoretical Physics, University of Cambridge}
\affil[13]{Institute of Astronomy, University of Cambridge}
\affil[14]{Kavli Institute for Cosmology, University of Cambridge}
\affil[15]{Department of Physics, University of Wisconsin-Madison}
\affil[16]{Universit\'e Grenoble Alpes, CNRS, Grenoble INP, LJK}
\affil[17]{Department of Chemistry, New York University}
\affil[18]{Department of Astrophysical Sciences, Princeton University}
\affil[19]{Lewis-Sigler Institute for Integrative Genomics, Princeton University\vspace{3pt}}
\affil[20]{Department of Physics, New York University}

\date{\vspace{-1em}\small $^*$Equal contribution\quad $^\dagger$Core contributors}

\begin{document}

\maketitle

\begingroup
\nnfootnote{Corresponding authors: Siavash Golkar, \texttt{siavash.golkar@gmail.com}; Alisha N. Jones, \texttt{alisha.jones@nyu.edu}; Olga Troyanskaya, \texttt{ogt@princeton.edu}; Shirley Ho, \texttt{shirleyho@flatironinstitute.org}}
\endgroup

\vspace{-3.5em}
\begin{abstract}
Biological function emerges from coupled constraints across sequence, structure, regulation, evolution, and cellular context, yet most foundation models in biology are trained within one modality or for a fixed forward task. We present MIMIC, a generative multimodal foundation model trained on our newly curated and aligned dataset, LORE, linking nucleic acid, protein, evolutionary, structural, regulatory, and semantic/contextual modalities within partially observed biomolecular states. MIMIC uses a split-track encoder-decoder architecture to condition on arbitrary subsets of observed modalities and reconstruct or generate missing components of molecular state across the genome, transcriptome, and proteome. Multimodal conditioning consistently improves MIMIC's sequence reconstruction relative to sequence-only inputs, while its learned representations enable state-of-the-art performance on RNA and protein downstream tasks. MIMIC achieves state-of-the-art splicing prediction, and its joint generative formulation enables isoform-aware inference that further improves performance. It accurately predicts nucleotide-specific DMS reactivity and guides RNA secondary structure prediction, outperforming conventional minimum free-energy (MFE) folding. Beyond prediction, the same generative framework supports constrained design. For RNA, MIMIC identifies corrective edits in a clinically relevant HBB splice-disrupting mutation without reverting it by using evolutionary and structural signals. In vitro splicing assays further validate MIMIC’s ability to design sequences that restore wild-type splicing while retaining the pathogenic mutation, consistent with RNA structural context and established splice-site scoring. For proteins, jointly conditioning on shape and surface chemistry of PD-L1 and hACE2 binding sites produces diverse, high-confidence sequences with strong in silico support for target binding. Finally, MIMIC uses experimental context as semantic conditioning to model assay-dependent RNA chemical probing, rather than treating context as a fixed output. Together, these results position MIMIC’s aligned multimodal generative modeling as a strong foundation for unifying representation learning, conditional prediction, and constrained biomolecular design within a single model.

\end{abstract}

\clearpage



\section{Introduction}\label{sec:intro}

The central dogma of molecular biology is often visualized as a linear, deterministic flow of information \cite{Crick1970CentralDogma}. However, the biological reality is a probabilistic web of feedback loops where downstream functions constrain upstream sequences, and environmental contexts dictate phenotypic outcomes. The function of a gene is not solely encoded in its coding sequence and promoter but is modulated by a shifting landscape of epigenetics and trans-acting factors \cite{Roadmap2015Epigenomes,ENCODE2020Expanded,Buenrostro2013ATAC}; similarly, while the amino acid sequence of a protein carries much of the information needed to specify its fold \cite{Anfinsen1973ProteinFolding} both folding and function are realized within a crowded cellular milieu shaped by chaperones and quality-control systems \cite{Hartl2011Chaperones}. Evolution further couples these layers: measures of evolutionary constraint reflect selection over joint molecular states rather than any single modality in isolation \cite{Siepel2005PhastCons,Pollard2010PhyloP,Davydov2010GERP}. To truly understand a biological system, one must model not just the forward flow of information, but the full joint probability distribution of its molecular states.


Despite this interdependence, the current landscape of artificial intelligence in biology remains fragmented. We have witnessed a proliferation of ``siloed experts''—models that master a single molecular modality but lack the capacity to operate across the broader central dogma. By focusing almost exclusively on forward prediction (mapping a sequence to a specific phenotype), these models fail to capture the full joint distribution of biological states. Consequently, they leave complex inverse problems unsolved: for example, given a desired protein structure, mRNA stability, and splicing pattern, what upstream nucleotide sequence is most likely to have produced them? This limitation is widespread, characterizing large language models trained on biomedical  \cite{lee2020biobert,tu2024towards, fallahpour2025bioreasonincentivizingmultimodalbiological, protrex, pubmedbert,chaves2024txllmlargelanguagemodel, Luo_2022, labrak2024biomistral}, protein sequences \cite{bepler2019learning, alley2019unified,rives2021biological,esm2,xtrimopglm,elnaggar2021prottrans,truong2023poet,Hayes2025ESM3,truong2025understanding,su2025trimodal,xu2023protst}, or nucleic acid \cite{Brixi2025Evo2,Boshar2025NucleotideTransformerV3,Nguyen2023HyenaDNA,Schiff2024Caduceus,Zhou2023DNABERT2} sequences. It is equally prevalent in forward regulatory genomic models \cite{Avsec2021Enformer,Avsec2026AlphaGenome,Kelley2018Basenji,Zhou2015DeepSEA,Chen2022Sei,Linder2025Borzoi,Kelley2020CrossSpecies}, structure prediction systems for both RNA \cite{yu2024interpretable,calonaci2020machine,shen2024accurate,townshend2021geometric,justyna2023machine} and proteins \cite{Jumper2021AlphaFold2,Baek2021RoseTTAFold,Abramson2024AlphaFold3,wu2022high}, and narrowly scoped phenotype predictors such as those for splicing \cite{Jaganathan2019SpliceAI,Zeng2022Pangolin,litman2026variant}.

Constructing a framework to capture this complex joint distribution has previously been difficult not simply because biology is multimodal, but because its measurements are heterogeneous in form, scale, and provenance: sequences, structures, evolutionary signals, and functional assays have different data types, are distributed across disparate resources, and are represented in incompatible coordinate systems. Different biological properties are only sparsely measured together; thus most samples will feature only a subset of possible modalities, requiring an architecture that can model observed data and infer missing components. These challenges are compounded by the range of context lengths required to capture both regulatory and biomolecular dependencies, which can span from local amino acid environments to kilobase-scale genomic neighborhoods. Existing biomolecular foundation models typically comprise a single-stream architecture, rendering them poorly suited to handle the task of learning across the joint distribution of heterogeneous biological data. Recent ``omnimodal'' scientific architectures instead treat heterogeneous measurements as first-class tokens via modality-specific tokenization \cite{4M,Parker2025AION1}, then learn a unified latent representation across diverse modalities via an encoder-decoder architecture. Our key insight is that by leveraging a flexible omnimodal architecture adapted to handle the heterogeneous context lengths and alignment structures of biological data, a \textbf{single} unified model can approximate the joint distribution of molecular states across the central dogma.

\begin{figure}[htbp]
    \centering
    \includegraphics[width=0.85\textwidth]{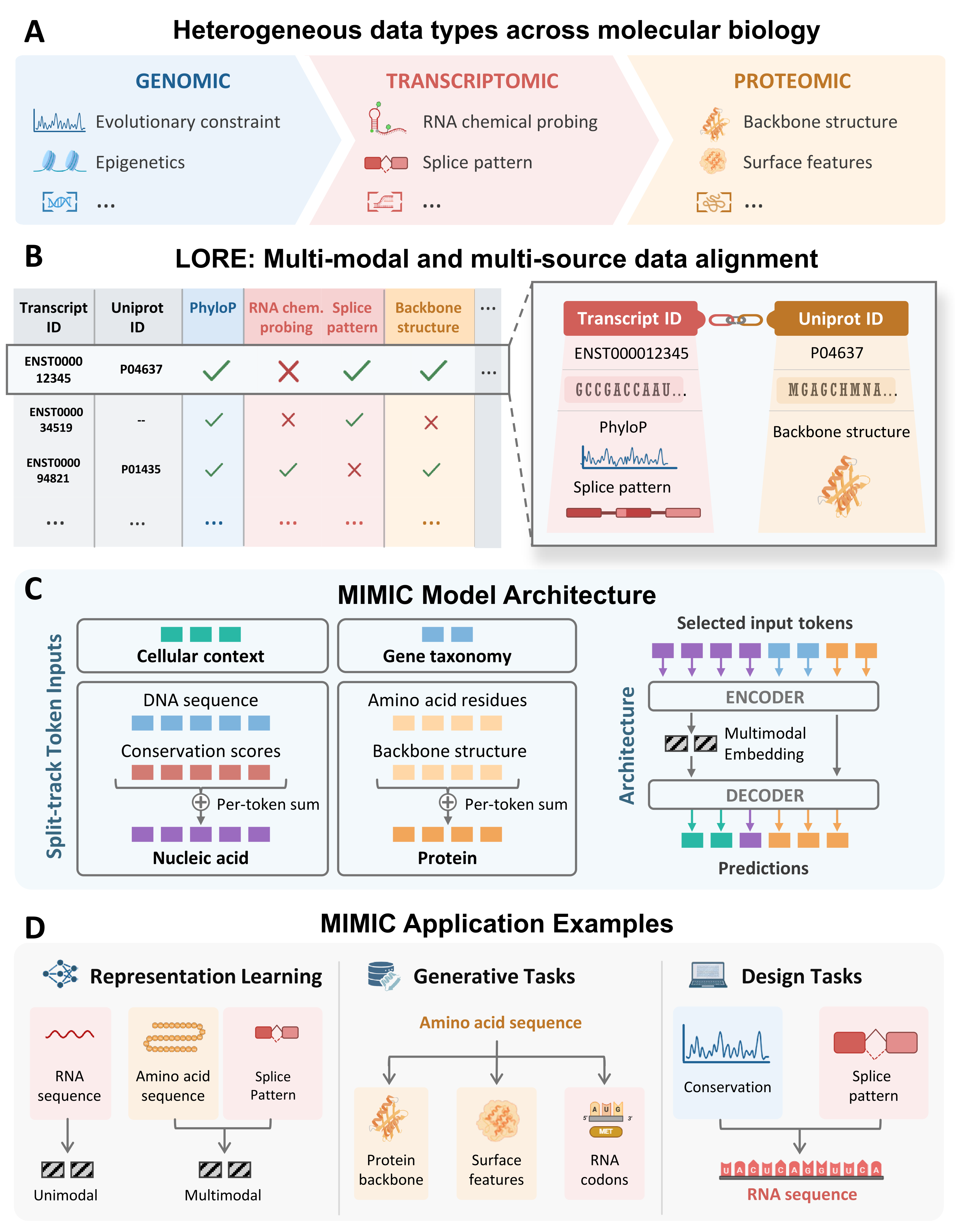}
    \caption{\footnotesize\textbf{Overview of the MIMIC framework.} \textbf{(A)} Molecular biology data is highly heterogeneous, spanning genomic, transcriptomic, and proteomic sequences, each with multiple assays and measurements that describe their function. \textbf{(B)} We curate LORE, a multimodal dataset that integrates and aligns data from multiple repositories into a set of unified but partially observed training examples. \textbf{(C)} Using this data, we build MIMIC, a multimodal foundation model for heterogeneous molecular biology. MIMIC models these examples using a split-track encoder-decoder architecture that combines co-aligned within-track signals while preserving distinct semantic inputs. \textbf{(D)} MIMIC enables multimodal representation learning, highly accurate property and splicing prediction, and constrained generative design across molecular biology.}
    \label{fig:hero}
\end{figure}

We introduce \textbf{MIMIC}, a generative foundation model with a novel split-track architecture designed to unify biomolecular data. MIMIC learns to translate fluently between the languages of the genome, the transcriptome, and the proteome while also interfacing with context-aware text representations that emphasize cellular specificity (\cref{fig:hero}a). Rather than solely predicting a quantity of interest or a single-track effect from sequence alone, MIMIC can process a rich molecular phenotype (e.g., uniting mRNA sequences with splice patterns, chemical reactivity scores, and a description of experimental conditions) to make \textit{conditioned} predictions, including sequence edits and designs (\cref{fig:hero}c). This capability provides a rich opportunity to interrogate the interdependence between modalities across the central dogma of biology, offering new avenues for discovery of regulatory mechanisms, protein design, and the interpretation of non-coding variants in the context of human health (\cref{fig:hero}d). A key obstacle to this kind of modeling has been the lack of aligned multimodal training objects: biological measurements are distributed across incompatible resources, are unevenly observed, and rarely co-occur in a form that allows direct learning over shared molecular state. To address these limitations, we introduce \textbf{LORE}, a curated dataset released alongside MIMIC. LORE aligns heterogeneous, multi-scale biological data into unified per-entity snapshots, providing the critical cross-modal supervision necessary for joint distribution training (\cref{fig:hero}b). Together, MIMIC and LORE move biological foundation modeling beyond siloed predictors and toward a holistic computational framework. 


In this work, we demonstrate that MIMIC is a state-of-the-art foundation model for predicting biomolecular properties, outperforming specialized models for proteins, nucleic acids, variant prediction, and splice site prediction. We show that multimodal conditioning reduces the search space of molecular design, enabling generation of high-quality proteins from a desired molecular surface and recovery of wild-type splicing patterns in the presence of pathogenic mutation. By integrating natural language descriptions of cellular and experimental context directly into the latent space, MIMIC shifts the traditional modeling paradigm from rigid, track-specific output toward flexible semantic conditioning; we find that context-conditioned predictions of chemical probing scores improve the computational modeling of RNA 2D structures. Our results highlight that MIMIC captures base-specific chemical reactivity and improves RNA secondary structure prediction over conventional MFE-based approaches. We further demonstrate the functional relevance of MIMIC through in vitro splicing assays, demonstrating its ability to design sequences that rescue aberrant splicing without reverting the pathogenic mutation, in agreement with RNA structural context and established splice-site scoring. Despite operating at a comparable size to other single-modality foundation models (1 billion parameters), MIMIC demonstrates the value of diverse and heterogeneous data as an alternative to pure scaling for improving model performance. This work represents a step toward a unified computational model of cellular biomolecules: one capable of learning bidirectionally across the molecular layers of the central dogma, and of translating directly between primary sequences and biological outcomes.

\section{MIMIC: A multimodal foundation model for molecular biology}
\label{sec:model}

To address the structural and contextual complexity of the central dogma, we developed MIMIC: a generative foundation model designed to jointly represent and reason over the heterogeneous molecular data of the central dogma. Because biological measurements span fundamentally different coordinate systems from nucleotide transcripts to amino acid sequences, MIMIC employs a novel split-track encoder-decoder transformer architecture. This design organizes biological inputs into distinct track groups, each suited to the coordinate system and molecule of the underlying modality, and learns a unified latent representation across them. We train a 1 billion parameter version of MIMIC, (see \cref{tab:mimic_architecture_params} for architecture details) on sequences up to 10,000 tokens using this split-track tokenization scheme. The encoder processes a unified token sequence assembled from several distinct track groups: a nucleic acid track, an amino acid track, and separate token groups for semantic and contextual information, alongside a set of learnable register tokens that aggregate global state across all tracks (\cref{app:framework}).

Within the nucleic acid and amino acid tracks, features sharing a common coordinate system (for example, phyloP conservation scores aligned to nucleotide positions, or DSSP secondary structure labels aligned to residue positions) are each embedded and summed element-wise into a single track representation. This split-track summation keeps total sequence length tractable regardless of how many modalities are present, while preserving the positional alignment between co-localized signals. Unaligned global information, including natural language descriptions of cellular context and functional annotations, occupies separate token groups concatenated to the biological tracks. Positional encoding uses Rotary Position Embeddings (RoPE) with a local group-reset strategy in which the position index restarts at zero for each track group, so that attention patterns reflect within-track distances rather than absolute token offsets across the full sequence. A cross-attention decoder with a 1,000-token context window queries the encoder's latent representation to reconstruct or generate arbitrary target modalities. This decoupling of observed context from predicted outputs is central to MIMIC; the asymmetric design (a wide receptive field for context, a focused window for generation) allows MIMIC to condition on any observed subset of modalities and generate any target subset without architectural modification or task-specific output heads, enabling tasks as distinct as generating a protein sequence from surface geometry, inferring a splice pattern from transcript boundaries, or predicting a phyloP conservation profile from sequence context alone.

MIMIC is trained using a staged curriculum that progressively scales the encoder context window from 1,000 to 10,000 tokens, allowing the model to first learn local sequence features before resolving long-range regulatory and structural dependencies. To handle the combinatorial heterogeneity of available modalities across samples, training is organized into a set of approximately 25 heuristic pathways, each specifying a required and optional set of input and target modalities. This pathway sampling ensures that rare but high-value combinations (for example, jointly observed structure and splice data) are adequately represented during training relative to more abundant sequence-only samples. Register tokens are trained via a reconstruction objective in which random token dropout forces them to encode sufficient information for the decoder to recover masked inputs, encouraging the registers to function as compressed global state vectors rather than simple copy mechanisms (\cref{sec:training}).

\section{LORE: a massively multimodal aligned biomolecular dataset}
\label{sec:dataset}

Training a multimodal architecture like MIMIC requires a fundamental shift in how biological data is curated: observations cannot be siloed by modality, but must be explicitly aligned to reflect their biological interdependencies. To provide this critical infrastructure, we curated LORE, a massively multimodal dataset of genomic, transcriptomic, and proteomic measurements containing approximately 15.5 million proteins and 13 million RNA transcripts from over 6000 organisms spanning all domains of life, along with over 4 billion tokens of biomedical, functional, and contextual text (\cref{app:dataset}). The central design principle of LORE is \emph{alignment}; this design is distinct from massive sequence-only corpora such as OpenGenome2 \cite{ArcInstitute2025OpenGenome2} or BFD \cite{Steinegger2018Linclust} that do not provide aligned or multimodal samples. Each row of the LORE dataset is an observation of either a transcript, a protein, or a pairing of the two, containing the subset of the modalities that are available for these biomolecules. Rather than treating different biomolecules as disconnected observations, we organize them into unified samples so that measurements referring to the same underlying genomic entity can be modeled jointly in their natural coordinate system. Because available data are inherently incomplete, LORE is intentionally constructed as a \emph{partially observed} multimodal dataset: no example is required to contain all modalities. Instead, we preserve high-value overlapping subsets wherever alignment is available, while reducing redundancy through sequence clustering and representative selection \cite{Steinegger2017MMseqs2} (\cref{app:alignment}). At the nucleic acid level, LORE includes tracks aligned with the unspliced RNA sequence \cite{OLeary2016RefSeq,Mudge2025GENCODE} such as evolutionary conservation \cite{Pollard2010PhyloP}, RNA chemical probing \cite{Mu2025RASPv2}, and regulatory assays \cite{Kodzius2006CAGE,Buenrostro2013ATAC} (\cref{app:lore_nucleic}). At the protein level, it includes amino acid sequence \cite{UniProt2025}, structural \cite{Varadi2022AlphaFoldDB} and surface features \cite{Gainza2020MaSIF}, functional descriptions \cite{boutet2007uniprotkb}, and measurements of abundance \cite{huang2023paxdb} (\cref{app:lore_protein}). Crucially, LORE also contains observations with \textit{semantic context}; epigenetic tracks, chemical probing data, and abundance measurements are paired with textual descriptions of the relevant organism, tissue, cell line, or experimental condition (\ref{app:lore_text}). These aligned multimodal observations make LORE a uniquely comprehensive substrate for training models that must reason jointly across the molecular layers of the central dogma, rather than learning each layer in isolation.

\section{Evaluating MIMIC across diverse biomolecular tasks}
\label{sec:validation}

\subsection{Multimodal sequence completion}
\label{sec:in-painting}

\begin{figure}[ht!]
    \centering
    \includegraphics[width=\textwidth]{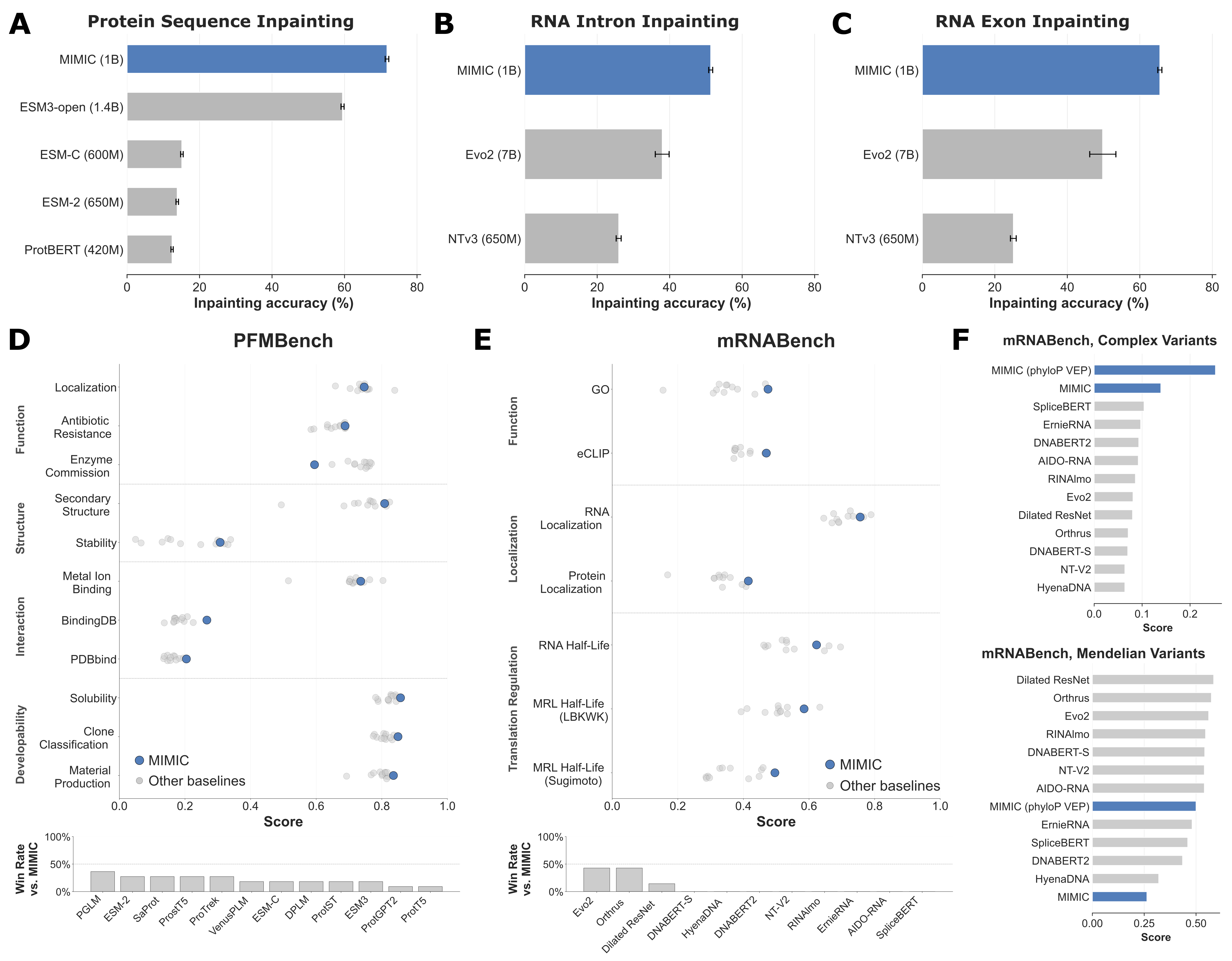}
    \caption{\footnotesize{\textbf{MIMIC achieves state-of-the-art performance across RNA and protein sequence property prediction benchmarks.} 
    \textbf{(A)} Per-residue top-1 amino acid inpainting accuracy at 100 masked positions. MIMIC (with structural and surface conditioning) outperforms all sequence-only protein language model baselines including ESM3-open, ESM-C, ESM-2 (650M), and ProtBERT. 
    \textbf{(B-C)} Per-nucleotide top-1 inpainting accuracy at 100 masked intronic (B) and exonic (C) positions. MIMIC (with RASP2, phyloP, and splicing conditioning) outperforms Evo 2 and NTv3 despite the latter two being evaluated on their training distribution (OpenGenome2).
    \textbf{(D)} PFMBench protein benchmark evaluation across function, structure, interaction, and developability tasks. MIMIC, even with sequence only input, is consistently among the top performing models. Below, we show the head-to-head win rate of all models against MIMIC; no model outperforms MIMIC on more than half the tasks.
    \textbf{(E)} mRNABench evaluation across RNA function, localization, and translation regulation tasks. MIMIC is the top performing model, outperforming Evo2 and Orthrus more than half the time and outperforming other models on nearly every task. 
    \textbf{(F)} On mRNABench complex variant effect prediction MIMIC is already the best performing method. Additionally, the phyloP VEP (\cref{sec:vep}) greatly outperforms all other methods. On Mendelian variants, phyloP VEP improves performance, but MIMIC does not surpass the best-performing baselines on this task.}}
    \label{fig:validation}
\end{figure}

Self-supervised pretraining is commonly instantiated as \emph{sequence completion}: predicting missing tokens from context. This objective underlies both masked and autoregressive language models, and therefore provides a natural starting point for comparing uni-modal baselines to a multimodal framework (\cref{fig:validation}a). To evaluate the performance of MIMIC over proteomic modalities, we likewise evaluate sequence completion on amino acid sequences, reporting per-residue reconstruction accuracy. We compare against several sequence-only protein language models (ProtBERT, ESM-2, ESM-C, and ESM3-open); MIMIC conditioned with structural descriptors achieves the highest reconstruction accuracy (\cref{fig:validation}a). To evaluate whether MIMIC has learned the distribution over genomic modalities, we formulated a sequence completion task on unspliced transcripts from MIMIC's test set, masking 100 nucleotides and reporting per-nucleotide top-1 accuracy at masked positions, stratified by exonic and intronic regions. We compared MIMIC with the full set of aligned genomic modalities against two long-context genomic foundation models trained on OpenGenome2: NTv3 \cite{Boshar2025NucleotideTransformerV3} (masked reconstruction) and Evo~2 \cite{Brixi2025Evo2} (autoregressive). MIMIC likewise achieves the highest reconstruction accuracy in both intronic (\cref{fig:validation}b) and exonic (\cref{fig:validation}c) regions. To clarify the contribution of multimodal conditioning, we ablated subsets of aligned auxiliary modalities (\cref{app:seq-completion}). Even in the sequence-only configuration, MIMIC is competitive with the top sequence-only models, and conditioning on more modalities leads to consistent improvements to sequence recovery. We find that multimodal conditioning is especially useful in cases where additional modalities are ambiguous rather than easily predictable from the primary sequence.

\subsection{Predicting downstream RNA and protein properties}
\label{sec:downstream}

One of the most powerful uses of biological foundation models is their ability to enable transfer learning to downstream tasks \cite{sledzieski2024democratizing,schmirler2024fine,hu2022lora}. We next investigated whether MIMIC's learned representations were informative for predicting several downstream properties of both nucleic acid and amino acid sequences, using two foundation model benchmarking suites: PFMBench \cite{pfmbench} and mRNABench \cite{mrnabench}. Following the protocol for each benchmark, we trained a supervised probe on MIMIC embeddings for each task (\cref{app:pfmbench}, \cref{app:mrnabench}). PFMBench \cite{pfmbench} covers protein function, structure, interaction, and developability tasks (\cref{fig:validation}d). Across these categories, MIMIC consistently matches or exceeds strong protein-only baselines including ESM3 \cite{Hayes2025ESM3}, ESM-C \cite{esm-c}, ProTrek \cite{protrex}, and SaProt \cite{saprot} (\cref{fig:validation}d). We show the head-to-head win rate for MIMIC against each of the models evaluated in the benchmark. MIMIC outperforms each other model on at least $7/11$ tasks. MIMIC achieves especially strong performance on the protein-ligand binding tasks (BindingDB, PDBbind), suggesting that jointly pretraining on surface chemical properties improves protein representations for predicting binding. MIMIC is also the best performing method for all developability tasks.

We next evaluated MIMIC on mRNABench \cite{mrnabench}, which includes several tasks covering prediction of mRNA function, localization, transcriptional regulation, and variant effect prediction (\cref{fig:validation}e, f). Here too, MIMIC is a state-of-the-art RNA foundation model. MIMIC outperforms Evo 2 \cite{Brixi2025Evo2} and Orthrus \cite{fradkin2025orthrus} on 4/7 of the tasks, Dilated ResNet \cite{mrnabench} on 6/7 tasks, and outperforms all other methods on all tasks. MIMIC displays the strongest relative performance for tasks that benefit from training on aligned protein modalities, such as GO function prediction ($+2\%$ vs. Evo2, $+9\%$ vs. Orthrus) and protein localization prediction ($+2\%$ vs. Evo2, $+5\%$ vs. Orthrus). Among the translation-related tasks in mRNABench, MIMIC shows its clearest gain on the Sugimoto half-life data set \cite{sugimoto2022isoform}, with a relative improvement of $8\%$ (vs. Orthrus). Because the Sugimoto data measures mean ribosome load for endogenous human transcript isoforms, this result suggests that MIMIC captures features relevant to isoform-level translational output. In contrast, MIMIC does not surpass the previous best model on the LBKWK half-life data set \cite{leppek2022combinatorial} ($-8\%$ vs. Orthrus), a benchmark derived from synthetic mRNA constructs in which stability-related effects are likely to play a larger role. These results suggest that MIMIC is particularly strong for endogenous transcript-level translation prediction. We note that with 566M encoder parameters (only the encoder is used to generate embeddings), MIMIC achieves state-of-the-art performance while being one of the smaller networks evaluated here.

\subsection{Variant Effect Prediction with MIMIC-predicted constraint}
\label{sec:vep}

For a sequence-only foundation model, downstream use is necessarily mediated through its latent representation: in the absence of an explicit aligned output modality, task transfer must proceed through probing or fine-tuning. In contrast, MIMIC can expose task-relevant signal directly through generation itself, rather than forcing every downstream task to be learned from a generic embedding. Variant effect prediction (VEP) provides a clear example. We report the performance of MIMIC in the mRNAbench VEP tasks under two inference scenarios. In the first (denoted as MIMIC in \cref{fig:validation}f), we follow the mRNAbench protocol to train a linear probe over the latent state. In the second, denoted as ``MIMIC  (phyloP VEP)'', we use the model's wild-type phyloP prediction together with a smoothed mutant-wild-type phyloP difference and train a logistic regressor on top of this two dimensional feature space (\cref{app:vep}). The two signals are complementary: reference-site conservation is most informative in regions already under detectable evolutionary constraint, whereas the mutation-induced signal remains informative in weakly conserved sequence, where pathogenicity depends more directly on mutation-specific disruption of local regulatory logic. This extremely low-dimensional regressor is more interpretable than representation-based alternatives and performs substantially better in practice than the embedding-based approach (\cref{fig:validation}f). On complex variants, the embedding-based version of MIMIC's VEP is already the strongest ($+33\%$ vs. SpliceBERT \cite{chen2024self}, the next-best model), but the phyloP-based VEP vastly outperforms all embedding-based approaches ($+82\%$ vs. embedding-based MIMIC). On Mendelian variants, switching from embedding-based to phyloP-based VEP improves MIMIC performance by 90\%.

\section{From splicing prediction to splice-conditioned RNA design}
\label{sec:splice}

\begin{figure*}[htbp]
    \centering
    \includegraphics[width=\linewidth]{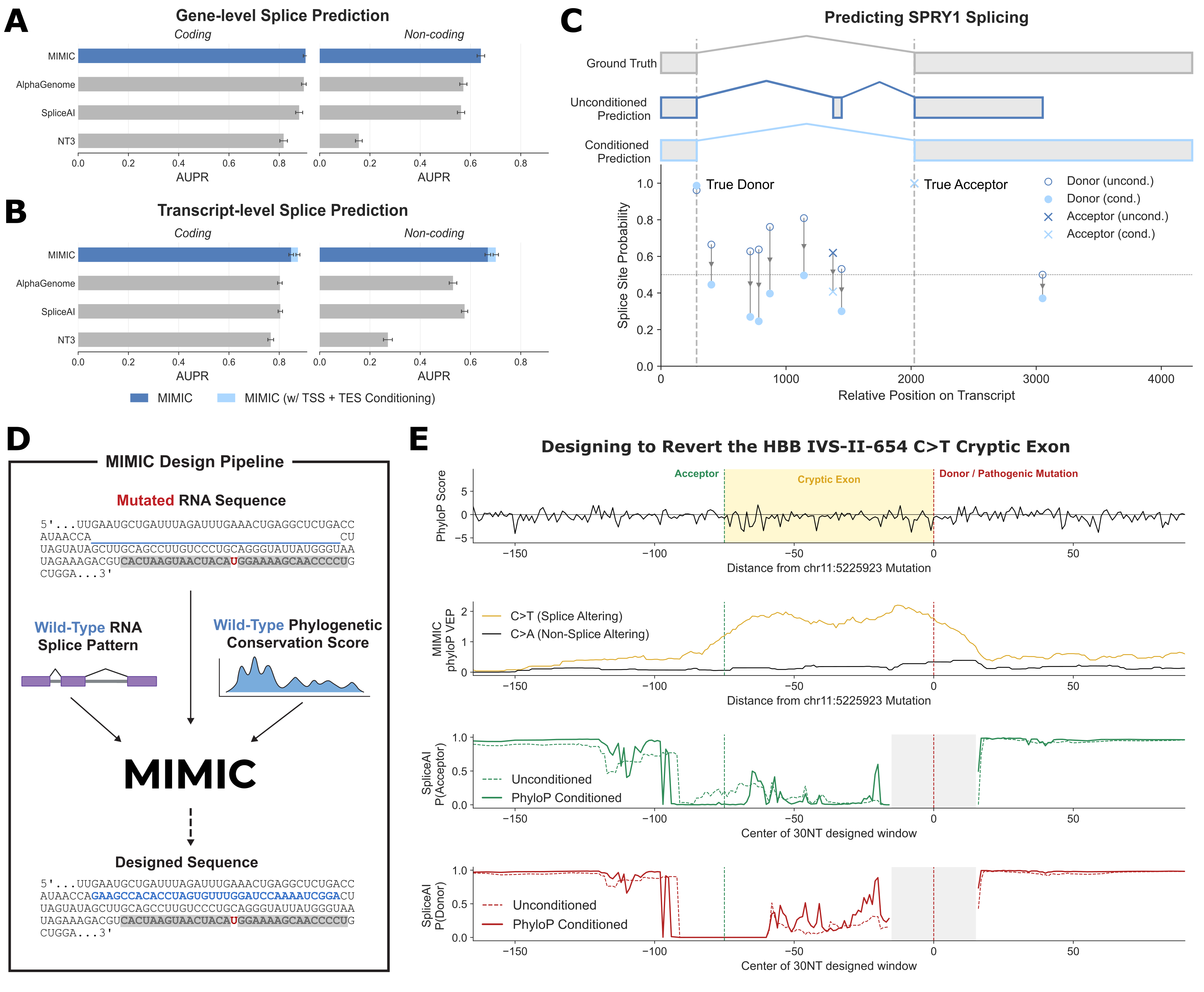}
    \caption{\footnotesize{\textbf{MIMIC accurately predicts splice sites and designs RNA sequences with predictable splice patterns.}
\textbf{(A)} Gene-level splice site prediction: MIMIC takes a genomic region as input and predicts donor and acceptor positions. Across coding (left) and non-coding (right) regions, MIMIC outperforms AlphaGenome, SpliceAI, and NTv3.
\textbf{(B)} Transcript-conditioned splice prediction: providing transcript context in terms of transcription start (TSS) and end (TES) sites enables transcript-specific inference. MIMIC sees larger gains in performance over previous models when conditioning on TSS and TES.
\textbf{(C)} An example of splice prediction with and without transcription specific conditioning. Without TSS and TES conditioning, MIMIC predicts an additional exon inclusion in \textit{SPRY1}. After conditioning, the correct splice architecture is predicted. The predicted probability of several false positive acceptor and donor sites in \textit{SPRY1} is substantially reduced following TSS and TES conditioning, leading to prediction of only the true splice junctions.
\textbf{(D)} Starting from a mutant sequence with the pathogenic variant fixed, MIMIC conditions on the wild-type splice pattern and wild-type phyloP to re-design a nucleotide sequence that suppresses inclusion of the cryptic exon.
\textbf{(E)} We apply this design pipeline to the \textit{HBB} IVS-II-654 $C>T$ mutation. (Top.) Wild-type conservation score, annotated with the cryptic acceptor and donor sites induced by the mutation. (Middle.) MIMIC phyloP VEP localized around the cryptic exon changes significantly between the wild-type sequence and the splice-altering $C>T$ variant. In contrast, a the non-splice-altering $C>A$ control shows no difference in predicted conservation from the wild type. (Bottom.) We show the predicted probability of pathogenic acceptor or donor sites for 30-nucleotide windows designed by MIMIC. MIMIC designs are significantly less likely to result in the inclusion of the cryptic exon. To avoid circularity, we use SpliceAI as an independent oracle for acceptor/donor probability at the cryptic sites.}}
    \label{fig:splicing}
\end{figure*}

\subsection{Predicting transcript-specific splice patterns}
\label{sec:splice-inference}

Alternative splicing is a central mechanism of transcript diversity and gene regulation, and its disruption is a major contributor to human disease \cite{Rogalska2023SplicingPhysiologyDiseaseTherapeutics,Jaganathan2019SpliceAI, litman2026variant}. To evaluate MIMIC as a splice site predictor, we conditioned the model on DNA sequence alone and asked it to predict donor (5') and acceptor (3') splice sites across a held-out set of human genes (\cref{fig:splicing}a). We compare with SpliceAI \cite{Strauch2022CISpliceAI}, AlphaGenome \cite{Avsec2026AlphaGenome}, and NTv3 \cite{Boshar2025NucleotideTransformerV3}, and report performance stratified by coding and non-coding genes as all models perform significantly better in coding regions. MIMIC outperforms these models (\cref{fig:splicing}a, $+1\%$ AUPR vs. AlphaGenome, $+3\%$ vs. SpliceAI, $+11\%$ vs. NTv3 in coding regions; $+12\%$ vs. AlphaGenome, $+14\%$ vs. SpliceAI, $+311\%$ vs. NTv3 in non-coding regions), demonstrating that a generative multimodal model can match dedicated forward predictors and genomic-only foundation models on classical splice-site detection.

However, per-site evaluation offers an incomplete picture of splicing mechanisms.
Prior work has shown that collapsing evaluation into per-site targets introduces noise into training and reduces clinically relevant predictive performance \cite{Strauch2022CISpliceAI}, and long-read transcriptomic studies have reinforced that biologically informative splicing patterns are only visible when transcript elements are considered jointly \cite{Belchikov2024LongReadIsoformExpression,Zhu2021DiseaseAssociatedRNAIsoforms,Joglekar2024BrainSpecializedSplicingPatterns,Inamo2024DiseaseLinkedIsoforms}. To circumvent this, MIMIC is jointly trained on full transcript realizations; it learns where splice sites occur within valid transcripts. 

We therefore additionally evaluated all models at the transcript level, prompting MIMIC to recover the full internal splice structure of held-out human isoforms. At this level of evaluation, MIMIC is clearly the strongest splicing model (\cref{fig:splicing}b, $+6\%$ AUPR vs. AlphaGenome, $+5\%$ vs. SpliceAI, $+11\%$ vs. NTv3 in coding regions; $+26\%$ vs. AlphaGenome, $+16\%$ vs. SpliceAI, $+146\%$ vs. NTv3 in non-coding regions). Furthermore, because MIMIC is a multimodal model, it can be conditioned explicitly on transcript boundary information. We show that explicitly providing the transcript start site (TSS) and transcript end site (TES) as input to MIMIC alongside the genomic sequence further improves recovery of internal splice structure ($+3\%$ in coding regions, $+5\%$ in non-coding regions, light blue in \cref{fig:splicing}b), consistent with evidence that transcription start site usage shapes downstream isoform selection \cite{AlfonsoGonzalez2024AlternativeTSSRNAProcessing,AlfonsoGonzalez2023TSSIsoformSelection}. As a representative example, we show that transcript conditioning improves MIMIC predictions of splicing in \textit{SPRY1} (\cref{fig:splicing}c). Prior to conditioning, MIMIC predicts an extra exon not included in the ground truth, as well as several alternative donor sites. After conditioning, all false positive splice junctions are predicted with significantly lower probability, so only the true exons remain. These results position MIMIC as a state-of-the-art splicing model that moves beyond single-locus prediction to global modeling of transcript architecture.

\subsection{Multimodal conditioned design of aberrant splicing mutations}
\label{sec:corrective_design_splice}

One of the central motivations for MIMIC is that a multimodal foundation model should support generative design tasks, not only forward prediction. Our results on sequence completion and splicing prediction suggest that MIMIC uses multimodal context to tighten the conditional distribution over biologically plausible sequences. We sought to investigate whether the same conditional machinery could be repurposed for an inverse problem: starting from a pathogenic regulatory context, can the model generate sequences that restore a wild-type-like state?

\paragraph{In silico design of compensatory edits.}
We first show that MIMIC preferentially regenerates the wild-type allele when the pathogenic position is masked, and that phyloP conditioning significantly improves this recovery, confirming that the model's conditional distribution is biased toward reference-compatible states (\cref{app:vep}). To move beyond this single-nucleotide demonstration, we next asked whether MIMIC could support a more demanding form of design: restoring wild-type function even when the pathogenic allele is held fixed. We investigated this in the intronic $\beta$-globin (\textit{HBB}) variant \textit{c.316-197C$>$T} (IVS-II-654 C$>$T), a canonical splice-altering mutation that promotes pseudoexon inclusion through aberrant splice-site gain \cite{Cheng1984, Lu2022} (\cref{fig:splicing}d). This setting provides a stringent test of whether the model can use multimodal context to identify compensatory sequence rewrites outside the causal locus, rather than merely regenerating the reference base. Furthermore, although wild-type phyloP conservation in this region provides no indication of the pathogenic nature of this mutation, MIMIC's phyloP VEP score clearly distinguishes the pathogenic C$>$T substitution from a matched non-splice-altering C$>$A control, localizing the signal to the cryptic acceptor and donor (\cref{fig:splicing}e).

We therefore prompted MIMIC with both the wild-type splice pattern and the pathogenic sequence with the C$>$T variant fixed (\cref{fig:splicing}e), generating designs for editable windows of 30 or 50 nucleotides centered at varying distances from the mutation (\cref{app:rna_splice_design}). Designed sequences were first evaluated in-silico using SpliceAI as an independent oracle to determine the probability of cryptic site gain at the pathogenic position.\footnote{AlphaGenome's splice prediction did not correctly predict the cryptic acceptor for the mutant sequence and was therefore not suitable for evaluating the corrective effect of MIMIC designs.} MIMIC produces designs that substantially reduce the likelihood of pathogenic splicing while maintaining the causal mutation. Furthermore, guided by the observation that deviation from wild-type phyloP is a marker of pathogenicity, we also prompted MIMIC to produce designs that recreate not only the wild-type splice pattern but also the wild-type phyloP profile. In this more demanding setting, MIMIC again generates designs that significantly and consistently reduce the likelihood of cryptic exon inclusion (\cref{fig:splicing}e). That effective corrections emerge even when the editable window does not directly overlap the cryptic splice sites demonstrates that multimodal conditioning supports indirect regulatory rewiring through the modification of splicing regulatory elements.

\begin{figure}[t]
    \centering
    \includegraphics[width=1\linewidth]{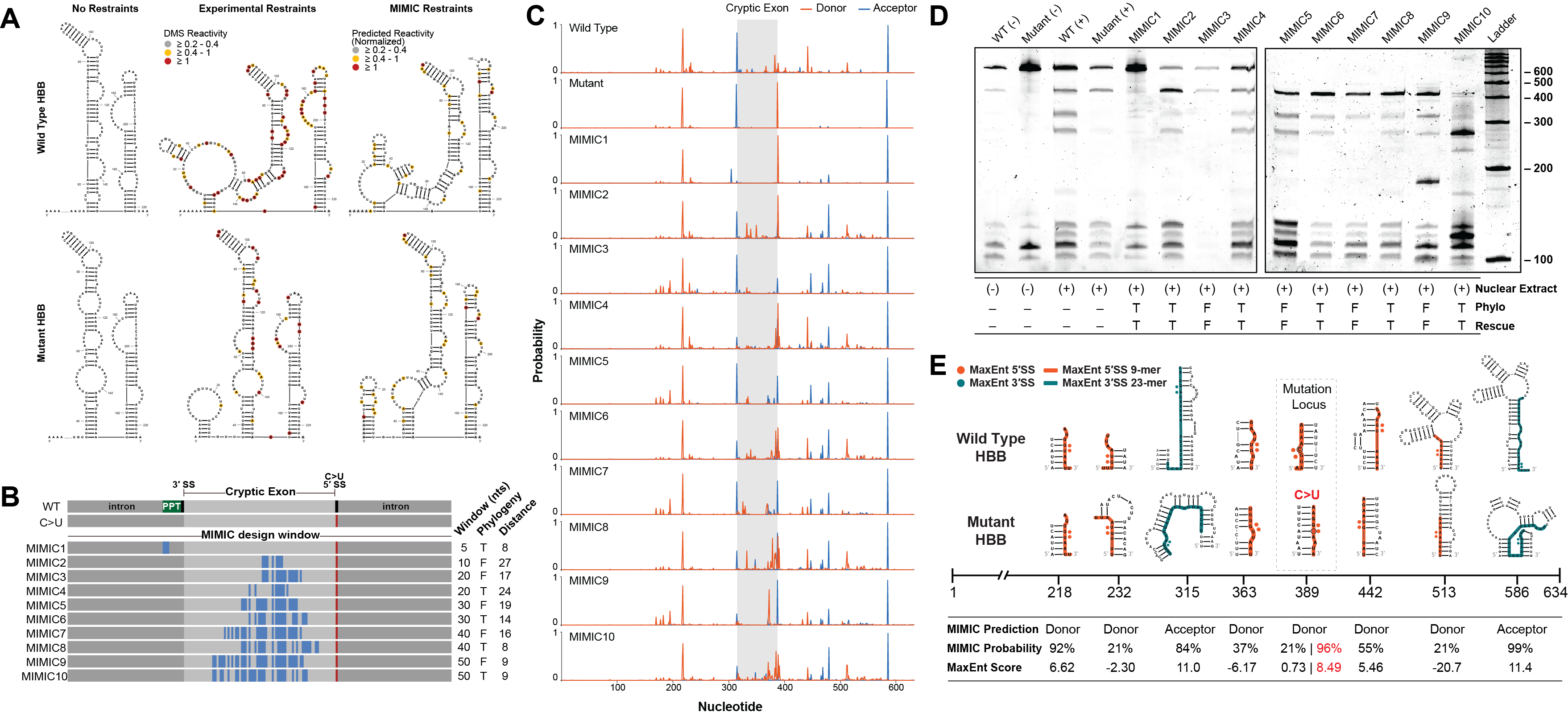}
    \caption{\footnotesize{\textbf{MIMIC accurately predicts secondary structure and splicing patterns of HBB minigene. (A)} Minimum free energy (MFE) secondary structures of wild type human beta-globin (HBB) and variant IVS-II-654 C>T (mutant HBB). Unconstrained MFE (left), dimethyl sulfate (DMS) probing restraints (middle), and MIMIC predicted DMS restraints (right).  \textbf{(B)} Minigenes for the wild type HBB, mutant HBB, and ten MIMIC designed constructs. The C>U mutated nucleotide is shown in red, nucleotides modified in the MIMIC designs shown in blue. PPT = polypyrimidine tract.  \textbf{(C)} MIMIC predicted probabilities of each nucleotide being a donor (orange) or acceptor (blue).  \textbf{(D)} PAGE gel analysis of PCR-amplified HBB minigene amplicons following in vitro splicing assays. Inclusion of phylogenetic context and whether rescue was successful are indicated. Constructs were considered rescues if bands of approximately 450 nts and 120 nts in size were recovered. \textbf{(E)} Splice-site motifs of the wild type and mutant HBB, along with their associated MIMIC predicted probabilities and MaxEnt scores. 5' splice sites are highlighted in orange and 3' splice sites are highlighted in blue. MIMIC probability and MaxEnt score for the mutant are shown in red.}}
    \label{fig:placeholder}
\end{figure}

\paragraph{Experimental validation of predicted \textit{HBB} RNA secondary structures.}
Because local RNA structure modulates splice-site accessibility, we next tested whether MIMIC accurately captures the secondary structure of the \textit{HBB} transcript. We in vitro transcribed and purified constructs representing the HBB wild-type (WT) and C>U mutant RNAs and chemically probed using dimethyl sulfate (DMS) (Figure 4a, Figure S8a). We then compared the reactivities of each nucleotide and resulting 2D structures to those predicted by MIMIC. Notably, the model correctly captured base-specific chemical reactivity, predicting DMS reactivity exclusively at adenosine and cytosine nucleotides while assigning no reactivity to guanosine or uracil. Folding the RNAs using the predicted reactivity profiles revealed that both the WT and C$>$U mutant transcripts are structured, with elevated reactivity localizing to single-stranded elements such as apical loops, bulges, and internal loops (Figure 4a).

To benchmark these structural predictions experimentally, we \textit{in vitro} transcribed and purified constructs representing the \textit{HBB} WT and C$>$U mutant RNAs and probed them with DMS \textit{in vitro} (Figure 4a, Figure S8a). Normalized reactivities predicted by MIMIC and measured experimentally showed a modest positive correlation for both constructs (Pearson $R = 0.24^*$ for WT and $R = 0.20$ for the C$>$U mutant; $^*P < 0.05$, Figure S8a). Lower nucleotide-level correlation is expected given that MIMIC was trained on cellularly transcribed RNAs, where intracellular pH, ionic conditions, RNA-binding proteins, temperature, and base modifications shape reactivity, and that predicted values in the generated \texttt{rasp2} track are smoothed to improve regional accuracy. Indeed, inspection of discordant positions revealed that disagreements occur predominantly within single-stranded elements rather than between paired and unpaired regions: MIMIC and experiment often assign peak reactivity to different nucleotides within the same apical loop or bulge. We therefore interpret MIMIC's structural resolution as operating at the level of secondary-structure elements rather than individual nucleotides, which is sufficient to recover helix boundaries via constrained folding. Consistent with this, the MIMIC-guided folds are more accurate than minimum free-energy (MFE) structures and closely match the 2D architectures derived from our in-house DMS probing (Figure 4a).

\paragraph{Experimental validation of corrective splicing designs.}
Finally, to test whether MIMIC's designed sequences restore normal splicing in a biochemical setting, we \textit{in vitro} transcribed and purified truncated minigene constructs for the WT, C>U mutant, and ten MIMIC-designed variants (Figure 4b, Figure S8b, c). Each construct contained 315 nucleotides of native 5' intron, the cryptic exon, 196 nucleotides of downstream native intron, and the first 50 nucleotides of exon 3, enough sequence for the secondary structural elements that dictate splice-site accessibility to fold (Figure S8b,c). As these constructs do not contain the native exon 2 5' splice site, we expected their splicing patterns to differ from those of the full length mRNA. Consistent with this, MIMIC predicted a new donor site 98 nucleotides downstream of the cryptic exon in addition to the cryptic 3' and 5' splice sites of the pseudoexon (Figure 4c). Following incubation in splicing-competent cellular extract, RT-PCR revealed distinct band profiles for the WT and C>U mutant RNAs, confirming that the cell-free assay recapitulates mutation-induced aberrant splicing (Figure 4d, Figure S8d). For the WT RNA, MIMIC predicted multiple candidate donor and acceptor motifs upstream of the cryptic 5' splice site, whereas the C>U mutation greatly increased the probability of splicing only at the cryptic site (Figure 4c). All MIMIC designs introduced new acceptor and donor sites, and the designs generated with phylogenetic information placed donor sites around the cryptic 5' splice site, resembling the WT profile. We see this represented in our gels: the mutant favored a single product and the wild type and MIMIC designs resulted in multiple bands. All five phylogeny-informed designs produced band patterns that closely resemble WT, demonstrating restoration of WT-like splicing behavior despite retention of the pathogenic mutation. 

Notably, the WT and C>U mutant RNAs share identical splice signal sequences except at the 5' splice site, yet they produce drastically different splicing patterns. Secondary structure analysis around the candidate donor and acceptor motifs showed that splice site accessibility differed at many splice junctions, indicating the point mutation substantially changes RNA structure (Figure 4e). We compared the MIMIC probability of each site with scores from MaxEntScan and saw a close relationship, with MIMIC probabilities above 50\% corresponding to high MaxEnt scores. We conclude that motif scoring alone cannot pinpoint exact junction usage without accounting for local RNA structure and the trans-acting factor repertoire present in the extract (Figure 4e, Figure S8b). Together, these data show MIMIC can design sequences that restore WT splicing patterns while retaining the pathogenic mutation, and its predictions are consistent with RNA structural context and established splice site scoring.

\begin{figure}[t!]
    \centering
    \includegraphics[width=\linewidth]{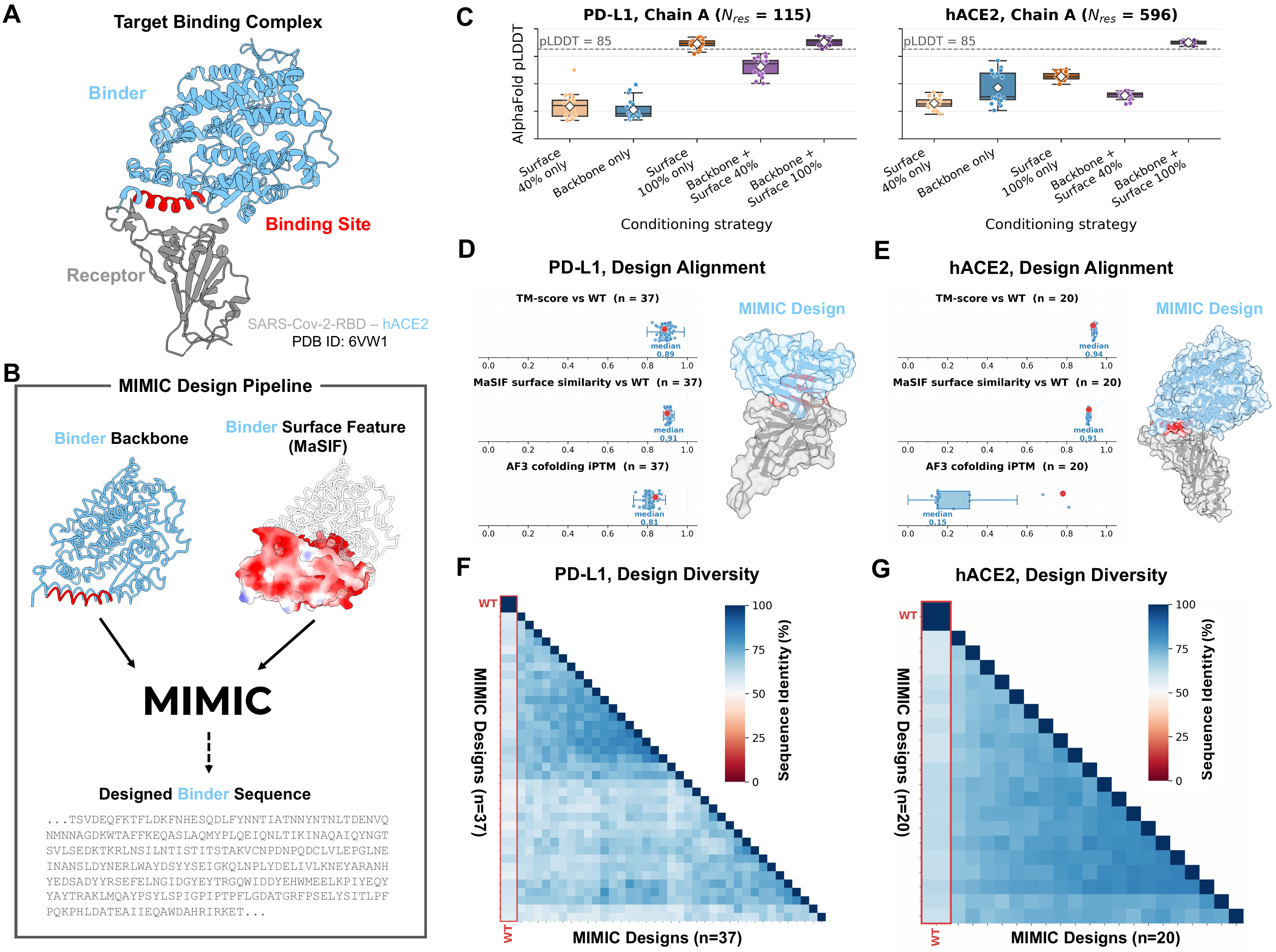}
    \caption{\footnotesize{\textbf{MIMIC designs recover target-binder properties with high sequence diversity.} \textbf{(A)} Schematic of the target binding complex (binder in blue, receptor in grey, binding site in red), use SARS-Cov-2-RBD - hACE2 (PDB ID: 6VW1) as an example. \textbf{(B)} Overview of the MIMIC design pipeline. The model generates novel sequences conditioned on the wild-type (WT) binder's backbone coordinates, MaSIF surface fingerprints, or both. \textbf{(C)} AlphaFold2 confidence (pLDDT) of designs across five conditioning strategies for a small PD-L1 (PDB ID: 4ZQK) and large hACE2 (PDB ID: 6VW1) target. Diamonds indicate the mean; the dashed line indicates the pLDDT = 85 quality threshold.
    \textbf{(D, E)} \textit{In silico} evaluation of high-confidence designs (pLDDT $>$ 85). Metrics shown are backbone TM-score versus WT (top), MaSIF surface similarity (middle), and AlphaFold3 cofolding iPTM (bottom). Red dots highlight the exemplar designs shown in the structural insets.
    \textbf{(F, G)} Pairwise sequence identity matrices for high-confidence designs and the WT sequence (red, top-left), ordered by hierarchical clustering on sequence dissimilarity.}}
    \label{fig:prot_design}
\end{figure}

\section{Multimodal conditioned design of structured binders}
\label{sec:protein-design}

Having demonstrated that multimodal conditioning improves our ability to generate nucleotide sequences with specific regulatory properties, we reasoned that an analogous approach could reduce the search space within a protein design framework. Our approach here builds directly upon the rich literature of structure- and surface-centric design \cite{gainza2020deciphering,gainza2023novo,wang2022scaffolding,Dauparas2022ProteinMPNN,Watson2023RFdiffusion}. Rather than treating backbone scaffolding and surface complementarity as separate design tasks, MIMIC's joint sequence-structure-surface representation allows simultaneous conditioning on both, explicitly encoding the physical, chemical, and electrostatic environment that mediates protein-protein interactions.

To test this, we targeted two therapeutically relevant proteins with well-characterized binding interfaces: Programmed death-ligand 1(PD-L1, PDB: 4ZQK), a major checkpoint immunotherapy target that engages Programmed Cell Death Protein 1 (PD-1) to suppress T-cell activation, and human Angiotensin-converting enzyme 2 (hACE2, PDB: 6VW1), the host receptor for the SARS-CoV-2 spike protein. For each target, we generated sequence ensembles under five conditioning strategies (20 per strategy totaling 100 designs each): MaSIF surface features alone (covering 40\% or 100\% of residues proximal to the binding site), backbone geometry alone, or backbone combined with partial or full surface features (\cref{fig:prot_design}a). We evaluated design quality using the AlphaFold2 predicted local distance difference test (pLDDT). Across both targets, design confidence reached the maximum when both backbone structure and surface chemistry were provided. Notably, the impact of multimodal conditioning scaled with target complexity: while providing the surface alone was sufficient to achieve high-confidence designs (pLDDT $\ge$ 85) for the smaller PD-L1 target, the larger hACE2 target relied much more heavily on additional backbone conditioning to reach comparable confidence (\cref{fig:prot_design}b).

To assess structural fidelity and sequence diversity among high-confidence designs (pLDDT $\ge 85$; PD-L1: $n=37$, hACE2: $n=20$), we evaluated their backbone recovery, surface similarity, and predicted binding competence relative to the wild-type (WT) structures (\cref{fig:prot_design}d, e). Across both targets, the designed sequences consistently recovered the WT fold (median TM-score $>0.85$) and surface chemistry (median MaSIF similarity $>0.90$). Furthermore, in silico co-folding with AlphaFold3 provided evidence of retained binding activity: for the smaller PD-L1 target, 35 out of 37 high-confidence designs achieved iPTM scores $>0.75$ (median $iPTM = 0.81$). The hACE2 target presented a more challenging design task due to its larger size and complexity; only two of the 20 designs with high-confidence fold were predicted to bind the SARS-CoV-2 spike RBD (AlphaFold3 $iPTM>0.75$). Crucially, MIMIC did not simply memorize or produce trivial variations of the WT sequence. Pairwise sequence identity matrices revealed that the generated designs maintain only about 50\% sequence identity to the WT (\cref{fig:prot_design}f, g, red boxes). While the designs exhibit local clustering (resulting in blocks of higher mutual sequence identity among themselves as they converge on shared functional motifs) the design ensembles remain globally distinct from the original WT sequence.

These results illustrate a central advantage of multimodal generative modeling for protein design: the ability to constrain generative search using complementary structural and chemical priors. By simultaneously conditioning on both backbone geometry and molecular surface features, MIMIC efficiently narrows the vast sequence space toward an ensemble of diverse solutions that are structurally stable, chemically faithful, and highly compatible with the target interaction, all without collapsing into trivial wild-type memorization. This targeted reduction in design freedom, analogous to the transcript-boundary constraints demonstrated earlier for RNA splicing, suggests that MIMIC has internalized a sequence fitness landscape tightly aligned with the true biological and physicochemical constraints of the interaction interface.

\begin{figure}[!htb]
    \centering
    \includegraphics[width=\linewidth]{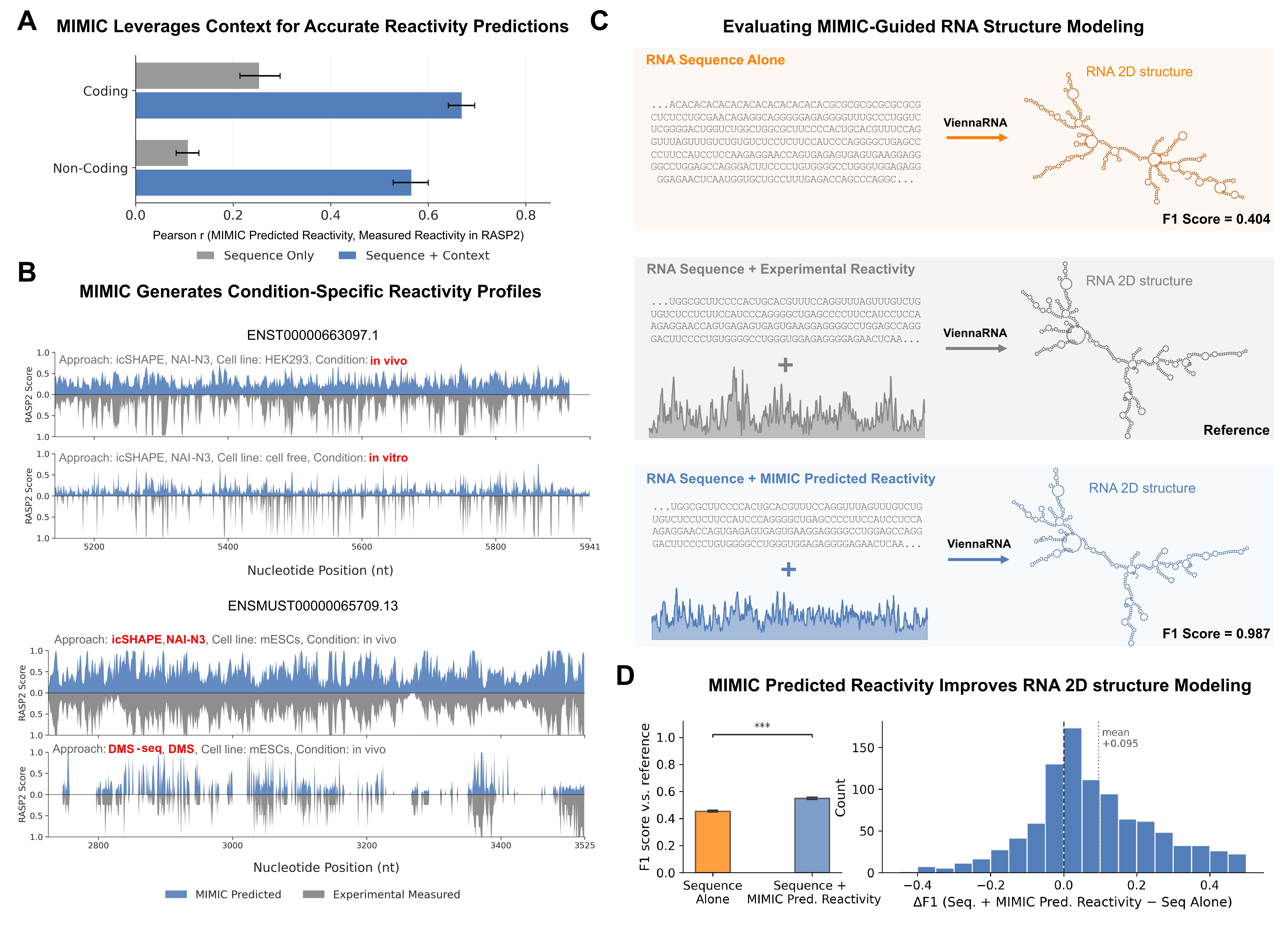}
    \caption{\footnotesize{\textbf{MIMIC leverages experimental context for accurate RNA reactivity prediction and RNA structure modeling.}} \textbf{(A-B)} MIMIC accurately predicts transcriptome-wide chemical probing reactivity (RASP2 scores) and adapts to condition-specific contexts. \textbf{(A)} Pearson correlation coefficients ($r$) between MIMIC-predicted and experimentally measured RASP2 scores for coding and non-coding RNAs. The context-aware generation (blue) significantly outperforms the sequence-only baseline (grey). Error bars represent standard error of the mean. \textbf{(B)} Representative reactivity tracks illustrating the model's sensitivity to cellular environment and probing approach. Top: MIMIC captures structural differences between \textit{in vivo} and \textit{in vitro} (cell-free) conditions for transcript \texttt{ENST00000663097.1}. Bottom: Predictions for transcript \texttt{ENSMUST0000065709.13} demonstrate the model's ability to differentiate between chemical probes, capturing the broad sensitivity of icSHAPE compared to the A/C-restricted sensitivity of DMS-seq. Blue tracks represent MIMIC predictions; grey tracks represent experimental measurements.\textbf{(C-D)} MIMIC-predicted reactivity improves RNA 2D structure modeling. \textbf{(C)} Structure modeling workflow using ViennaRNA evaluated on an 800 nt window of transcript \texttt{ENSMUST00000128851.8}. Incorporating MIMIC-predicted reactivity (blue, $\mathrm{F} 1=0.987$) yields a 2D structure nearly identical to the experimentally-guided reference (grey), vastly outperforming sequence-alone folding (orange, $\mathrm{F} 1=0.404$). \textbf{(D)} Quantitative evaluation of structural improvement across the test set ($\mathrm{n}=1,000$ RNA sequences). Left: Box plot comparing F1 scores of sequence-alone modeling versus MIMIC-guided modeling (*** $P=6.25 \times 10^{-47}$, one-sided paired Wilcoxon signed-rank test). Error bars represent standard error of the mean. Right: Histogram of the change in F1 score ($\Delta \mathrm{F} 1$) after incorporating MIMIC predictions. The rightward shift demonstrates robust structural improvement across the dataset (median $\Delta \mathrm{F} 1=+0.061$, mean $=+0.095 \pm 0.194$ s.d.).}
    \label{fig:rasp_context}
\end{figure}

\section{Context-conditioned RNA reactivity prediction to guide structure modeling}
\label{sec:rasp-context}

Biological processes are inherently dynamic, governed as much by their surrounding cellular environment and physiological state as by their underlying molecular sequence. Consequently, as we aggregate large-scale experimental datasets to train foundation models, we must recognize that these measurements do not exist in a vacuum. They are inextricably linked to the specific biological context, cellular state, and probing approach used during data collection. To faithfully capture this complexity, MIMIC is designed to model experimental conditions explicitly. Rather than relying on fixed output heads for different assays, MIMIC processes natural language descriptions of experimental conditions directly alongside molecular data. By incorporating these semantic inputs into its latent representation, the model learns to treat experimental context not as generic auxiliary metadata, but as a primary modality that dictates how a molecular state should be realized. Crucially, we demonstrate that this context-aware capability translates directly into practical benefits for downstream analytical tasks. By accurately predicting condition-specific reactivity profiles, MIMIC provides high-fidelity constraints that improve computational RNA 2D structure modeling.

We grounded this evaluation in chemical probing data from the RASP2 (RNA Atlas of Structure Probing) database\cite{Mu2025RASPv2}. Because chemical probing serves as the primary experimental workhorse for investigating RNA architecture, it provides an ideal and highly stringent computational test case; reactivity measurements for identical transcripts vary substantially depending on the probing approach and the cellular environment. We found that conditioning MIMIC on the correct natural language experimental context significantly improves transcriptome-wide RASP2 predictions relative to models conditioned on sequence alone (Figure \ref{fig:rasp_context}a). This performance gap demonstrates that MIMIC successfully learns to leverage semantic context to resolve biological ambiguities that sequence alone cannot address.

To confirm that these improvements reflect biologically meaningful adaptations rather than statistical artifacts, we inspected individual MIMIC predictions against experimentally measured RASP2 reactivity scores across diverse contexts (Figure \ref{fig:rasp_context}b). The model demonstrates a high degree of sensitivity to environmental variables; for example, it successfully captures the distinct structural profiles of the same transcript probed \textit{in vivo} versus \textit{in vitro} (cell-free). Furthermore, the model accurately adjusts its output based on the specified probing approach. When comparing an \textit{in vivo} icSHAPE assay (click selective 2-hydroxyl acylation and profiling experiment, carried out with the NAI-N3 probe) to dimethyl sulfate sequencing (DMS-seq), MIMIC successfully differentiates between the broad sensitivity of icSHAPE and the A/C-restricted sensitivity of DMS, generating the appropriate condition-specific reactivity tracks. This contextual conditioning extends beyond experimental parameters. We found that MIMIC can also leverage other biological modalities to uncover structural insights. For instance, conditioning the model on evolutionary conservation scores (PhyloP) yields a distinct uplift in RASP2 prediction accuracy (\cref{fig:phylop_rasp_supp}). This predictive synergy highlights an underlying biological correlation between RNA structure and evolutionary conservation, conceptually analogous to how coevolutionary information derived from multiple sequence alignments (MSAs) drives structural predictions in the protein folding field\cite{Jumper2021AlphaFold2}.

Beyond predicting reactivity profiles, we investigated whether MIMIC’s context-aware predictions provide tangible utility for downstream biological tasks. Specifically, we evaluated the use of MIMIC-predicted reactivity scores to guide computational RNA 2D structure modeling (Figure \ref{fig:rasp_context}c, details in Appendix \ref{app:shape_folding}). Integrating MIMIC predictions into standard folding algorithms, ViennaRNA\cite{lorenz2011viennarna}, generates plausible RNA secondary structures that are qualitatively nearly identical to structures guided by experimental reference data, vastly outperforming sequence-alone folding. Quantitatively, incorporating MIMIC-predicted reactivity profiles robustly improves structural modeling accuracy across a representative population of the test set (Figure \ref{fig:rasp_context}d). The resulting folds exhibit a significant increase in F1 scores compared to sequence-alone baselines.

Together, these results demonstrate that MIMIC successfully internalizes diverse biological modalities-ranging from semantic experimental descriptions to evolutionary conservation—as integral components of biological modeling. By leveraging these contextual inputs to specify the appropriate realization of a molecular measurement, MIMIC provides a powerful framework for generating and interpreting complex biological data, ultimately driving more accurate modeling and analysis across a wide array of downstream tasks.

\section{Discussion}\label{sec:discussion}

This study introduces MIMIC, a generative multimodal foundation model for heterogeneous biological data.
Across the results presented here, a consistent pattern emerges: aligned auxiliary context improves model representations and performance, consistent with prior findings from ``omnimodal'' scientific models \cite{Parker2025AION1}. To enable the training of MIMIC, we developed LORE, an aligned data set spanning diverse biomolecular modalities. We show that multimodal conditioning enhances sequence completion, yields transferable representations for RNA and protein downstream tasks, and supports transcript-level modeling of splicing. We apply MIMIC for multimodal biomolecular design, providing stabilizing constraints that increase design efficiency in both nucleic acid and protein settings, including experimentally validated rescue of aberrant splicing. MIMIC supports semantic context by modeling it as a conditioning variable within the same framework, rather than treating it as metadata, imposing a fixed output schema, or requiring separate track-specific outputs. Sequence, structure, regulation, evolutionary constraint, and experimental context can be understood as distinct but related observations of an underlying molecular state. This perspective helps explain why a single model can support tasks that are often studied separately, and our results indicate that biological modeling may benefit from frameworks that move flexibly between representation learning, conditional prediction, and constrained generation within a shared model.

Several limitations should, however, guide the interpretation of these findings. First, LORE captures only a subset of the measurements that shape molecular phenotype. Many important facets of molecular biology remain absent or only weakly represented. Even among the modalities that are included, coverage is highly uneven, so MIMIC is trained on a partial and imbalanced view of molecular biology, which may limit its performance in some applications relative to specialized models designed for specific tasks. In parallel, the current context scale remains limited relative to the biological processes one would ultimately wish to capture, and the restricted decoder window further limits the extent of direct generation over long outputs. A natural next step is to broaden modality coverage while improving alignment quality and metadata standardization. Particularly important additions include long-range chromatin and regulatory measurements, richer transcript-resolved assays, and more complete measurements of biomolecular function and interaction. Additionally, longer and explicitly multiscale architectures may be needed to connect local biochemical constraints with cell-state-scale regulation. On the conditioning side, semantic context would likely benefit from moving beyond sparse free text toward more structured representations of experimental state that still interface naturally with language.

Similar analytical caveats apply to our experimental validation of splice-site rescue. While the empirical RT-PCR banding patterns closely align with our \textit{in silico} combinatorial predictions, gel electrophoresis alone cannot definitively resolve which specific donor--acceptor pairs are utilized within each isoform. Sequence-based splice-site scoring provides an estimate of intrinsic motif strength, but actual splice-site selection is heavily modulated by local RNA secondary structure and the repertoire of \textit{trans}-acting RNA-binding proteins (such as SR proteins and hnRNPs) active within the cellular extract. Consequently, a strong consensus motif sequestered within a stable stem-loop may be bypassed in favor of a weaker, structurally accessible site, or vice versa, depending on factor availability. Thus, while the restoration of WT-like band profiles in six of the ten variants strongly demonstrates functional rescue of the splicing defect, direct Sanger or targeted amplicon sequencing of individual excised gel bands will be required in future work to map exact exon--intron junctions at nucleotide resolution.


We show that MIMIC unifies heterogeneous molecular measurements within a single generative framework, enabling missing components to be inferred from available context and allowing prediction and design to be framed as different forms of inference over the same underlying biomolecular system. More broadly, jointly modeling the interconnected layers of molecular biology through flexible multimodal representations of aligned molecular state represents a powerful new abstraction for biological foundation models.

\clearpage
\paragraph{Contributions.}

The author contributions are summarized below. The order of the authors within each category is randomized.

\begin{itemize}
    \item \textbf{Project Lead:} Siavash Golkar
    \item \textbf{Core Team:} Ksenia Sokolova, Jake Kovalic, Minhuan Li, Samuel Sledzieski, Alberto Bietti, Geraud Krawezik, Siavash Golkar, Irina Espejo Morales
    \item \textbf{Advisory}: Alisha N. Jones, Shirley Ho, Sonya M. Hanson, Olga G. Troyanskaya, Pilar Cossio
\end{itemize}

\subsubsection*{MIMIC Contributors} 
\begin{itemize}
    \item \textbf{Model architecture:} Jake Kovalic, Siavash Golkar, Irina Espejo Morales, Fran\c{c}ois Lanusse Lanusse, Liam Parker
    \item \textbf{Pretraining:} Irina Espejo Morales, Siavash Golkar, Jake Kovalic, Geraud Krawezik 
    \item \textbf{Downstream evaluations}: Siavash Golkar, Minhuan Li, Jake Kovalic, Ksenia Sokolova, Irina Espejo Morales, Shengwei Xiong, Alisha Jones, Samuel Sledzieski
    \item \textbf{RNA design task:} Shengwei Xiong, Siavash Golkar, Alisha Jones
    \item \textbf{Protein design task:} Minhuan Li, Samuel Sledzieski, Siavash Golkar, Vikram Mulligan
    \item \textbf{RNA 2D structure modeling task:} Minhuan Li, Irina Espejo Morales, Siavash Golkar
    \item \textbf{Manuscript writing:} Samuel Sledzieski, Siavash Golkar, Minhuan Li, Ksenia Sokolova, Alisha Jones, Jake Kovalic, Claudia Skok Gibbs, Irina Espejo Morales
\end{itemize}

\subsubsection*{LORE Contributors} 
\begin{itemize}
    \item \textbf{Modality alignment and merging:} Jake Kovalic, Irina Espejo Morales, Siavash Golkar, Samuel Sledzieski
    \item \textbf{RNA chemical probing (RASP):} Minhuan Li, Alisha N. Jones, Roman Klypa
    \item \textbf{Epigenetics:} Ksenia Sokolova, Siavash Golkar, Claudia Skok Gibbs
    \item \textbf{Protein sequence, structure and surface features:} Samuel Sledzieski, Minhuan Li
    \item \textbf{Gene and transcript sequence and annotations:} Jake Kovalic
    \item \textbf{Text processing and biomedical corpus:} Irina Espejo Morales
\end{itemize}

\subsection*{Other}
This section includes contributions ranging from code review and problem solving, compute and HPC support, as well as discussions and feedback.

\begin{itemize}
    \item Payel Mukhopadhyay, Miles Cranmer, Rudy Morel, Lucas Meyer, Helen Qu, Jeff Shen, Tom Hehir, Hadi Sotoudeh, Kyunghyun Cho, Mariel Pettee, David Fouhey
\end{itemize}

\paragraph{Acknowledgments}

We would like to acknowledge the support of the Simons Foundation and of Schmidt Sciences. This work was supported in part by the AI2050 program at Schmidt Sciences (Grant G-25-70028). A.N.J. acknowledges support from NIH grant 1R35GM160278-01. Additionally, computations were run at facilities supported by the Scientific Computing Core at the Flatiron Institute. The Flatiron Institute is a division of the Simons Foundation. The authors thank Justin Kinney, Doug Renfrew, and Bargeen Turzo for helpful discussions, and Lucy Reading-Ikkanda and Chi-Dat Lam for assistance with figures. The authors also thank the lab of Kristin Lynch for assistance with in vitro splicing assays.

\paragraph{Data and Model Availability}

We are currently preparing MIMIC code and weights for public release. We are also preparing LORE for public release, including data clustering, splits, and pre-computed tokenization. Upon release, both the MIMIC model and the LORE data set will be available on the Polymathic AI GitHub (\href{https://github.com/PolymathicAI}{https://github.com/PolymathicAI}). 

\clearpage
\bibliography{biblio}
\bibliographystyle{unsrt}

\clearpage

\appendix

\setcounter{figure}{0}
\renewcommand{\thefigure}{S\arabic{figure}}

\setcounter{table}{0}
\renewcommand{\thetable}{S\arabic{table}}

\startcontents[appendices]
\printcontents[appendices]{}{0}{\setcounter{tocdepth}{2}\section*{Materials and Methods}}

\clearpage

\section{The LORE dataset}
\label{app:dataset}

\begin{table}[ht!]
\centering
\small
\caption{\textbf{LORE modality composition summary.} The number of samples for contextual modalities include sequence $\times$ context pairs. The number of sequences is reported after reduction of redundancy based on sequence similarity.}
\begin{tabular}{lr}
\toprule
\textbf{Modality} & \textbf{Samples} \\
\midrule
\midrule
\multicolumn{2}{c}{\textit{Nucleic Acid Track}}\\
Nucleic acid sequence & 12,967,153 \\
Transcript annotation & 12,967,153 \\
Splice junctions & 12,967,153 \\
Coding sequence mask & 12,967,153 \\
Coding sequence junctions & 11,906,378 \\
Evolutionary conservation (phyloP) & 518,480 \\
Promoter usage (CAGE) & 15,073,298 \\
Chromatin accessibility (ATAC-Seq) & 14,021,301 \\
RNA chemical probing (RASP2) & 1,644,404 \\ 

\midrule
\multicolumn{2}{c}{\textit{Protein Track}}\\
Amino acid sequence & 15,607,838 \\
Secondary structure (DSSP) & 15,607,838 \\
Tertiary structure (backbone) & 15,607,838 \\
Solvent accessible surface area (SASA) & 15,607,838 \\
Chemical surface (MaSIF) & 1,671,206 \\
Protein abundance & 1,803,028 \\
Functional captions & 176,626 \\
\midrule
\multicolumn{2}{c}{\textit{Semantic \& contextual modalities}} \\
Taxonomic classification & 12,967,153 \\
Biomedical corpus & 3,797,568 \\
\bottomrule
\end{tabular}
\end{table}

\subsection{Overview}

A significant challenge in multimodal biological modeling is the scarcity of aligned data. Public repositories such as GenBank \cite{GenBank2025Update}, UniProt \cite{UniProt2025}, and the PDB \cite{wwPDB2019} contain extensive data, but these resources are often disjoint. To train MIMIC, we curated \textbf{LORE}, a multimodal dataset that aligns heterogeneous signals into a unified coordinate system. The modality sources in LORE were to provide multi-view and context-specific information across genomic, transcriptomic and proteomic layers of the central dogma. In the following sections, we provide a detailed description of the data sources, preprocessing, alignment and curation efforts that were necessary to build LORE.

\subsection{Tokenization}

Unless specified otherwise in their individual sections, all modalities used one of the following tokenization strategies.

\paragraph{Character tokenization}
Sequences of discrete characters were tokenized at the character level, with a distinct token for each possible value (e.g. $\{A, C, G, U\}$ for nucleotide sequences) plus \texttt{pad}, \texttt{mask}, and \texttt{unknown} tokens. Boolean/mask tracks were considered a special case of character tokenization with a vocabulary of $\{0, 1\}$. Modalities with character tokenization include: nucleic acid sequence ($n=4$), splicing mask ($n=2$), splice junctions ($n=3$), coding sequence mask ($n=2$), coding sequence junctions ($n=3$), coding sequence codons ($n=64$), chromatin accessibility ($n=12$), amino acid sequence ($n=24$), secondary structure ($n=9$).

\paragraph{Continuous tokenization}
Several modalities are sequences of continuous values. In this case, we randomly sampled $~20,000$ training samples and computed percentile bins, which were saved with the tokenizer. New samples were assigned to one of $n=100$ bins and assigned the corresponding token. We also included \texttt{pad}, \texttt{mask}, and \texttt{unknown} tokens for missing values. Bin centers were also saved and used for detokenization. Modalities with continuous tokenization include: evolutionary conservation, RNA chemical probing, promoter usage, solvent accessible surface area,  MaSIF \texttt{n\_vertices}, MaSIF \texttt{si\_index}, MaSIF \texttt{charge}, MaSIF \texttt{hbond}, MaSIF \texttt{hydrophobicity}, protein abundance

\paragraph{Class tokenization}
A few modalities appy a single label to the whole sample, rather than containing a sequence. In this case the data were tokenized as a single token, corresponding to the number of label options, plus \texttt{pad}, \texttt{mask}, and \texttt{unknown} tokens. Modalities with class tokenization: transcript annotation ($n=7$), taxonomic classification (kingdom, $n=11$).

\paragraph{Text tokenization}
Free text modalities were tokenized using the pre-trained WordPiece \cite{song2021fastwordpiecetokenization} tokenizer from Biobert \cite{lee2020biobert} (HuggingFace: \texttt{dmis-lab/biobert-base-cased-v1.2}). Modalities with text tokenization include: contexts for \{RASP2, ATAC-Seq, CAGE, Protein Abundance\}, protein functional captions, biomedical corpus.  

\subsection{Nucleic acid track}
\label{app:lore_nucleic}

LORE integrates physicochemical, evolutionary, and semantic modalities alongside standard sequence-structure pairs. The nucleic acid track aligns the RNA/DNA sequence (including dynamic flanking regions) with structural and evolutionary priors.

\paragraph{Nucleotide sequence}
Along with full human and mouse genomes and accompanying annotations from GENCODE \cite{Mudge2025GENCODE}, we integreate nucleic acid data from over 6000 additional organisms from NCBI RefSeq \cite{OLeary2016RefSeq}. We selected common model organisms including \textit{Drosophila melanogaster} (fruit fly), \textit{Caenorhabditis elegans} (roundworm), and \textit{Saccharomyces cerevisiae} (yeast), as well as all genomes tagged as \texttt{complete reference genomes} by RefSeq \cite{OLeary2016RefSeq}. These collected organisms span all domains of life and contain over 25 million transcripts and 123 billion nucleotides. 

We limit LORE to transcripts of at most 10,000 base pairs, along with any aligned associated modalities such as splice patterns, experimental assays, corresponding protein sequence and structure (see \ref{app:alignment}). All transcripts are padded with genomic flanking regions (200 BP on either side for Bacteria/Archaea, 1000 BP upstream/300 BP downstream otherwise) so that important regulatory context regarding is available in our transcript-level data. 

\paragraph{Transcript Annotation}
To distinguish genomic elements, we included discrete classification labels for coding status (coding, non-coding) and feature type \cite{Mudge2025GENCODE,OLeary2016RefSeq}. Because these feature types are non-standard across the varied annotators hosted by RefSeq \cite{OLeary2016RefSeq}, we merge synonyms and significantly condense our vocabulary to produce a tractable subset: \texttt{\{lncRNA, miscRNA, protein\_coding, pseudogene, rRNA, tRNA, other\}}.

\paragraph{Nucleotide-Level Annotation (Splicing)}
We additionally provide annotation of functional regions within the genomic transcripts assembled above \cite{Mudge2025GENCODE,OLeary2016RefSeq}. Binary masks for coding sequence (CDS) and spliced exons are extracted directly from annotated genomes. Explicitly labeled untranslated regions (UTRs), start codons and stop codons from GENCODE \cite{Mudge2025GENCODE} are collected in a similar manner, but we do not currently use them for training MIMIC. We then construct splice junction tracks from the exon masks, separately labeling donors and acceptors to align with previous splice prediction work \cite{Jaganathan2019SpliceAI, Avsec2026AlphaGenome, Boshar2025NucleotideTransformerV3}. We combine donors, acceptors, transcription start sites (TSS) and transcription end sites (TES) into a single multiclass track, but we evaluate only acceptor and donor prediction when comparing. 

\paragraph{Transcript coding sequence}
In addition to the nucleotide-level tacks discussed above, for protein coding regions we provide a spliced RNA representation which is tokenized per 3-mer. Each codon is aligned to the associated amino acid sequence and is thus summed with other protein modalities in the split-track architecture discussed in \ref{app:split-track} below. 

\paragraph{Evolutionary conservation (phyloP)}
Conservation scores derived from phyloP to \cite{Pollard2010PhyloP} describe nucleotide-level evolutionary conservation by testing whether the substitution rate at each position deviates from neutral drift under a phylogenetic model. We utilized 100-way alignments for human sequences and 30-way alignments for mouse sequences to capture lineage-specific constraints \cite{UCSCPhyloP100wayHg38,UCSCPhyloP30wayMm9}.

\paragraph{Chemical probing (RASP2)}
Transcriptome-wide chemical probing data\cite{wang2021rna} curated in RASP2\cite{Mu2025RASPv2} provides probabilistic measures of RNA base-pairing accessibility. These assays serve as indicators of secondary structure, across various experimental contexts and can be used as a bias to in-silico RNA secondary structure prediction algorithms. Since the published chemical probing data in RASP2 is aligned to genome releases that may be different from what is published in RefSeq \cite{OLeary2016RefSeq}, we use the LiftOver tool \cite{ucsc} to correctly align experimental tracks to genomic sequence. 

\paragraph{Chromatin accessibility (ATAC-Seq)}

Chromatin accessibility (ATAC-seq \cite{Buenrostro2013ATAC}) data were obtained from the ENCODE Project \cite{ENCODE2020Expanded} portal by querying all released \texttt{narrowPeak} BED files associated with ATAC-seq experiments. For each file, we extracted and stored accompanying metadata, including genome assembly, biosample term, release information, and accession identifier, and standardized these into a unified metadata table. Human experiments from hg38 were retained directly, whereas mouse experiments from mm10 were lifted over to mm39 using CrossMap \cite{zhao2014crossmap}. Metadata were updated so that successfully converted mouse files were reassigned to mm39, and metadata tables were then split by species to maintain a consistent organization across downstream processing steps.

To generate continuous chromatin accessibility tracks suitable for transcript-level integration, \texttt{narrowPeak} files were converted to \texttt{bigWig} \cite{kent2010bigwig} format using the \texttt{signalValue} field provided by ENCODE. Within each experiment, peak signal values were discretized into ten quantile-bins, assigning integer values from $1$ to $10$ to accessible regions, where higher values corresponded to stronger accessibility relative to other peaks in the same experiment. Prior to \texttt{bigWig} generation, peaks were sorted, overlapping intervals were merged using the maximum signal across overlaps, and intervals were clipped to chromosome boundaries using the appropriate genome-specific chromosome sizes. We then extracted per-base ATAC signal across transcript coordinates for human and mouse transcripts. For each transcript and experiment, bases overlapping an accessible region retained their corresponding rank-binned accessibility value, bases within mappable regions but lacking a peak were assigned $0$ to indicate closed chromatin in that experiment, and bases that could not be mapped because of missing chromosomes or transcript coordinates extending beyond chromosome boundaries were recorded as missing values. This produced a transcript-by-experiment representation of chromatin accessibility that preserved experiment-specific signal without averaging across contexts. 

\paragraph{Promoter Usage (CAGE)}

Cap Analysis of Gene Expression (CAGE) \cite{Kodzius2006CAGE} peak data was downloaded from the FANTOM5 repository at RIKEN phase 2.6. We used TPM (Transcripts Per Million) normalized CAGE peak data for the following species and genome builds: Human: hg19, Mouse: mm9, Rat: rn6, Rhesus Macaque: rheMac8, Chicken: galGal5, Dog: canFam3. In the case where a different build was required, we mapped CAGE peak locations between genome builds using LiftOver. Metadata (cell type, replicate, sample ID) was extracted from raw CAGE data file column names. As it was unstructured, we used the following regular expressions to identify donors and biological replicate (if donor info was not found)

\begin{itemize}
    \item Donor: \verb!re.match w/ r"(.+?)[,\s]+(donor\d+|pool\d+|day\d+)$!
    \item Replicate: \verb!r'\b(biol\_rep[l]?\d+|tech\_rep\d+|donor\d+|pool\d+|donation\d+)\b'!
\end{itemize}

\subsection{Protein track}
\label{app:lore_protein}

LORE also contains multiple views on proteins, aligning the amino acid sequence with several structural and physicochemical properties, as well as descriptions of its function.

\paragraph{Amino acid sequence}
To ensure that we had matching structures for the protein sequences in our data set, we started with $\approx 200$M structures download from AlphaFoldDB \cite{Varadi2022AlphaFoldDB} version 4 and extracted amino acid sequences from these structures. The UniProt ID was used as the primary identifier for joining and mapping aligned modalities within the protein track.

\paragraph{Secondary structure}
Secondary structure predictions were made using the \texttt{mkdssp} package (\href{https://github.com/PDB-REDO/dssp}{https://github.com/PDB-REDO/dssp}). We use the canonical 8-class secondary structure plus an unknown/loop class: $\alpha$ helix (H), $3_{10}$ helix (G), $5$ helix (I), extended strand (E), $\beta$ bridge (B), turn (T), bend (S), coil (C), loop (L).

\paragraph{Tertiary structure}
We retained entries from AlphaFoldDB with $pLDDT > 70$. Tertiary structures were tokenized using the pretrained ESM3 structure tokenizer \cite{Hayes2025ESM3} with $4096$ classes based on local backbone geometry. The ESM3 structure tokenizer is a vector-quantized variational autoencoder (VQ-VAE) trained to reconstruct protein structures from AlphaFoldDB, ESMAtlas \cite{esm2}, and antibody-antigen complexes from the PDB.

\paragraph{Solvent accessible surface area}
The solvent accessible surface area was computed on the AlphaFoldDB structures using the Shrake-Rupley ``rolling-probe'' algorithm \cite{Shrake1973SASA}, as implemented in the \texttt{biotite} Python package (version 1.4) \cite{kunzmann2023biotite}.

\paragraph{MaSIF surface features}
We computed MaSIF (Molecular Surface Interaction Fingerprinting) features\cite{gainza2020deciphering} on a subset ($\sim 10\%$) of our protein structures due to compute limits (~50,000 cpu hours for 1.6M datasets), prioritizing both sequence and structural diversity. These geometric fingerprints encode the electrostatic potentials, hydrophobicity and curvature of the protein surface, explicitly teaching the model to recognize interaction hotspots. To ensure consistency with residue-level tokens from other modalities, we aggregate surface vertex features to the corresponding residue. Specifically, a vertex is mapped to a residue if it lies within 2.8 Å of any of its atoms, a threshold corresponding to the diameter of a solvent molecule as established in the literature\cite{grudman2023optimal}. Based on these proximity mappings, we construct a residue-level feature vector by computing the mean for shape index and hydrophobicity, and the sum for electrostatic charge and hydrogen bonding. Additionally, the total number of mapped vertices is recorded as a proxy for solvent-accessible surface area. Completely buried residues lacking any proximal surface vertices are assigned NaN values.

\paragraph{Abundance}
We included experimental measurements of protein abundance from PaxDB v5.0 \cite{huang2023paxdb} as a context-dependent protein modality. The \texttt{paxdb-uniprot-links-v5.0} file was used to map from PaxDB record IDs to UniProt IDs; any non-matching records were discarded. As abundance values are already normalized across different experimental conditions in PaxDB, we were able to collate records from different species into a single unified data set.

\paragraph{Functional text}
We aligned sequences with natural language descriptions, including functional protein captions and gene family descriptions. The functional annotations were obtained from Swiss-Prot \cite{UniProt2025} for the fields of \texttt{function}, \texttt{family} and \texttt{subcellular location}. Preprocessing for captions modality removed database and publication IDs from text.

\subsection{Semantic \& contextual modalities}
\label{app:lore_text}

\paragraph{Biomedical corpus} We curated a standalone text corpus derived from exclusively from biomedical literature (PubMed) \cite{Sayers2026NCBI}. While this corpus is not aligned to specific sequences, it is tokenized using the same vocabulary as the semantic tracks \texttt{WordPiece} \cite{song2021fastwordpiecetokenization}. This enables the model to transfer semantic reasoning learned from the broad literature to the specific gene and protein descriptions. Overall after tokenization the corpus amounts to 3,797,568 samples and 4 billion tokens.

\paragraph{Taxonomic classification} We provide each gene's taxonomic family as plain text conditioning to MIMIC. Rather than attempting to create a standardized vocabulary for more than 600 families in our final dataset, we allow our natural language module described above to process the taxonomic information semantically. 

\paragraph{Cell line, tissue, and experimental context}\label{app:context}

\begin{itemize}
    \item \textbf{RASP2:} The context fields for this modality are extracted from the RASP2 database \cite{Mu2025RASPv2} and are: \texttt{approach}, \texttt{reagent}, \texttt{condition}, \texttt{cell line} and \texttt{description}. The description field mainly contains a summary of the other fields, sometimes misplaced fields or extra information, this field always goes last in the context string to prevent the model from looking into the future. Table \ref{tab:rasp_contexts} shows all the field combinations for human (the condition is standardized here so that the table is readable, but condition has not been standardized in our LORE dataset).
    
    \item \textbf{CAGE:} In the CAGE peaks dataset, human represents 67.9\% of the dataset and mouse 32\% with 0.2\% of the samples having a null species. The descriptions contain information about the tissue, cell types, cell lines and other descriptive fields about donors such as age or sex, examples are shown in \cref{tab:cage_contexts}.

    \item \textbf{ATAC:} In the ATAC dataset, human represents 85\% and mouse 15\% approximately. Examples of sub-contexts are shown in \cref{tab:atac_contexts}.

    \item \textbf{Protein Abundance:} The contexts for this modality have a free text format with minimum processing directly form the dataset source. The type of information they contain are: organism, tissue, donor information, cell line or condition. Not all contexts include all of the later information. A subsample of contexts is shown in \cref{tab:prot_abund_ctx}.
    
\end{itemize}

\begin{table}[!ht]
\caption{\textbf{Context descriptions for human and mouse within the CAGE peaks dataset.} We provide sample counts for each species, as well as the number of sub-contexts and several randomly sampled example contexts.}
\centering
\small
\begin{tabular}{p{4.2cm} r r p{7.2cm}}
\toprule
Organism & Count & \# Sub-Contexts & Examples of Sub-Contexts \\
\midrule
\midrule
\textit{Homo sapiens} (human) & 32.8M & 488 & Heart, CD34+ stem cells, Light melanocytes, HTMMT mixed Mullerian tumor cell line, Hs 5.T leiomyosarcoma cell line \\ \hline
\textit{Mus musculus} (mouse) & 15.4M & 106 & Intestine, Medulla oblongata, Striatal neurons, Granule cells, Spiral ganglion neurons \\
\bottomrule
\end{tabular}

\label{tab:cage_contexts}
\end{table}

\begin{table}[!ht]
\caption{\textbf{Context descriptions for human and mouse within the ATAC-seq dataset.} We provide sample counts for each species, as well as the number of sub-contexts and several randomly sampled example contexts.}
\centering
\small
\begin{tabular}{p{4.2cm} r r p{7cm}}
\toprule
Organism & Count & \# Sub-Contexts & Examples of Sub-Contexts \\
\midrule
\midrule
\textit{Homo sapiens} (human) & 30,618,542 & 206 & A549, Gm18907, Hg02884, Mcf-7, Lung \\
\textit{Mus musculus} (mouse) & 5,381,573 & 36 & Regulatory T Cell, Dendritic Cell, Heart, Megakaryocyte, Stomach \\
\bottomrule
\end{tabular}

\label{tab:atac_contexts}
\end{table}

\begin{table}[!ht]
\centering
\small
\caption{\textbf{Context descriptions within the Protein Abundance dataset.} We provide the number of samples from the five top species species and 5 varied sub-context examples that are sampled randomly to illustrate diversity. This table is not exhaustive, as a total of 155 normalized species and 637 unique contexts were included.}
\begin{tabular}{p{4.2cm} r p{7.2cm}}
\toprule
Context & Count & Examples of Sub-context \\
\midrule
\midrule
\textit{Homo sapiens} (human) & 1,752,363  & liver, HEK293, esophagus mucosa, A431, vitreous humor \\
\textit{Mus musculus} (mouse) & 483,996 &  brain, kidney cortex, midbrain, placenta, spleen B cells \\
\textit{Arabidopsis thaliana} (thale cress) & 367,701 &  silique, leaf, flower bud, flower, 1 d, rosette, circadian \\
\textit{Glycine max} (soybean) & 99,396  & seed, matured seed, shoot, shoot, nitrogen-fed, nodule, nitrogen-fed \\
\textit{Danio rerio} (zebrafish) & 84,782  & whole organism, testis, embryo 120 h, embryo 48 hpf, whole body \\
\bottomrule
\end{tabular}
\label{tab:prot_abund_ctx}
\end{table}

\begin{table}[!ht]
\centering
\small
\caption{\textbf{Unique RASP2 context combinations.} We show combinations of \texttt{approach}, \texttt{reagent}, \texttt{condition}, and \texttt{cell line} in the full dataset of 118 human normed datasets. The \texttt{description} field is dropped from the table for readability reasons.}
\begin{tabular}{llll}
\toprule
\texttt{approach}& \texttt{reagent}& \texttt{condition}& \texttt{cell line} \\
\midrule
\midrule
\multirow{12}{*}{icSHAPE} & \multirow{12}{*}{NAI-N3}
    & \multirow{4}{*}{in vivo}       & H9 \\
 & &                                  & HELA \\
 & &                                  & HEPG2 \\
 & &                                  & K562 \\
\cmidrule(l){3-4}
 & & in vitro                         & - \\
 & & chromatin in vitro               & - \\
 & & chromatin in vivo                & - \\
 & & cytoplasm in vitro               & - \\
 & & cytoplasm in vivo                & - \\
 & & nucleoplasm in vitro             & - \\
 & & nucleoplasm in vivo              & - \\
\midrule
\multirow{2}{*}{icLASER} & \multirow{2}{*}{Naz-N3}
    & in vitro                        & - \\
 & & in vivo                          & - \\
\midrule
\multirow{4}{*}{smartSHAPE} & \multirow{4}{*}{NAI-N3}
    & in vivo 125ng                   & HEK293 \\
 & & in vivo 1ng                      & HEK293 \\
 & & in vivo 25ng                     & HEK293 \\
 & & in vivo 5ng                      & HEK293 \\
\midrule
\multirow{2}{*}{Keth-seq} & \multirow{2}{*}{N3-kethoxal}
    & kethoxal vs no-treat            & HELA \\
 & & kethoxal vs no-treat in vitro    & HELA \\
\midrule
\multirow{6}{*}{DMS-seq} & \multirow{6}{*}{DMS}
    & \multirow{2}{*}{denatured}      & FBL \\
 & &                                  & K562 \\
\cmidrule(l){3-4}
 & & \multirow{2}{*}{in vitro}        & FBL \\
 & &                                  & K562 \\
\cmidrule(l){3-4}
 & & \multirow{2}{*}{in vivo}         & FBL \\
 & &                                  & K562 \\
\midrule
DMS-MaPseq & DMS & in vivo      & - \\
\bottomrule
\end{tabular}
\label{tab:rasp_contexts}
\end{table}

\subsection{Cross-track alignment, clustering, and splits}\label{app:alignment}


\paragraph{Sequence Clustering \& Selection}
To mitigate redundancy and data leakage, we clustered the protein and RNA sequences based on sequence similarity using MMseqs2 version \\ \texttt{87e7103d289029dc3345f85ea9a4c4c6d6416e46} \cite{Steinegger2017MMseqs2}. We used \texttt{-c 0.8} and \texttt{--min-seq-id 0.3} or \texttt{--min-seq-id 0.7} to obtain clusters of varying levels of similarity. 30\% clusters were used for creating splits, while 70\% clusters were used to ensure diversity in the training data. We selected a single representative sample per cluster, resulting in a final dataset of approximately 15.5 million protein samples and 13 million RNA samples. We show the sizes of each cluster in \cref{tab:clusters}.

\paragraph{Alignment} We mapped UniProt IDs to their corresponding RefSeq/Ensembl IDs where possible \cite{UniProt2025,OLeary2016RefSeq,Mudge2025GENCODE} using the \texttt{idmapping.dat} file provided by UniProt release \texttt{2024\_04}. This yielded a core subset of approximately 2 million aligned transcript/protein pairs. For gene clusters in this aligned set, we included all documented splice isoforms to capture alternative splicing patterns \cite{Mudge2025GENCODE,OLeary2016RefSeq}.

\begin{table}[!ht]
\centering
\caption{\textbf{Protein and RNA clustering sizes used in the data selection pipeline.}}
\begin{tabular}{lrrr}
\toprule
\textbf{Modality} & \textbf{Identity} & \textbf{Unique Clusters} & \textbf{Total Members} \\
\midrule
\midrule
Protein & 30\% & 42,943,944 & 299,566,796 \\
Protein & 70\% & 111,588,657 & 299,566,796 \\
RNA     & 30\% & 10,861,904  & 25,001,445  \\
RNA     & 70\% & 11,804,593  & 25,001,445  \\
\bottomrule
\end{tabular}
\label{tab:clusters}
\end{table}

\paragraph{Data splits}
Train, validation, and test splits were performed at the cluster level for both protein and RNA sequences \cite{Steinegger2017MMseqs2}. This ensures that no sequence in the validation set shares significant similarity with the training set. All aligned modalities (e.g., structure, conservation, expression) associated with a given sequence were assigned to the same split as their parent sequence.

\section{MIMIC architecture}
\label{app:framework}

Standard transformer architectures, designed for single-stream text or protein data, are poorly suited to efficiently model heterogeneous multimodal data. MIMIC addresses this challenge via a novel ``Split-Track'' architecture that sums aligned modalities while concatenating distinct molecular entities.

\begin{figure}[ht]
    \centering
    \includegraphics[width=\linewidth]{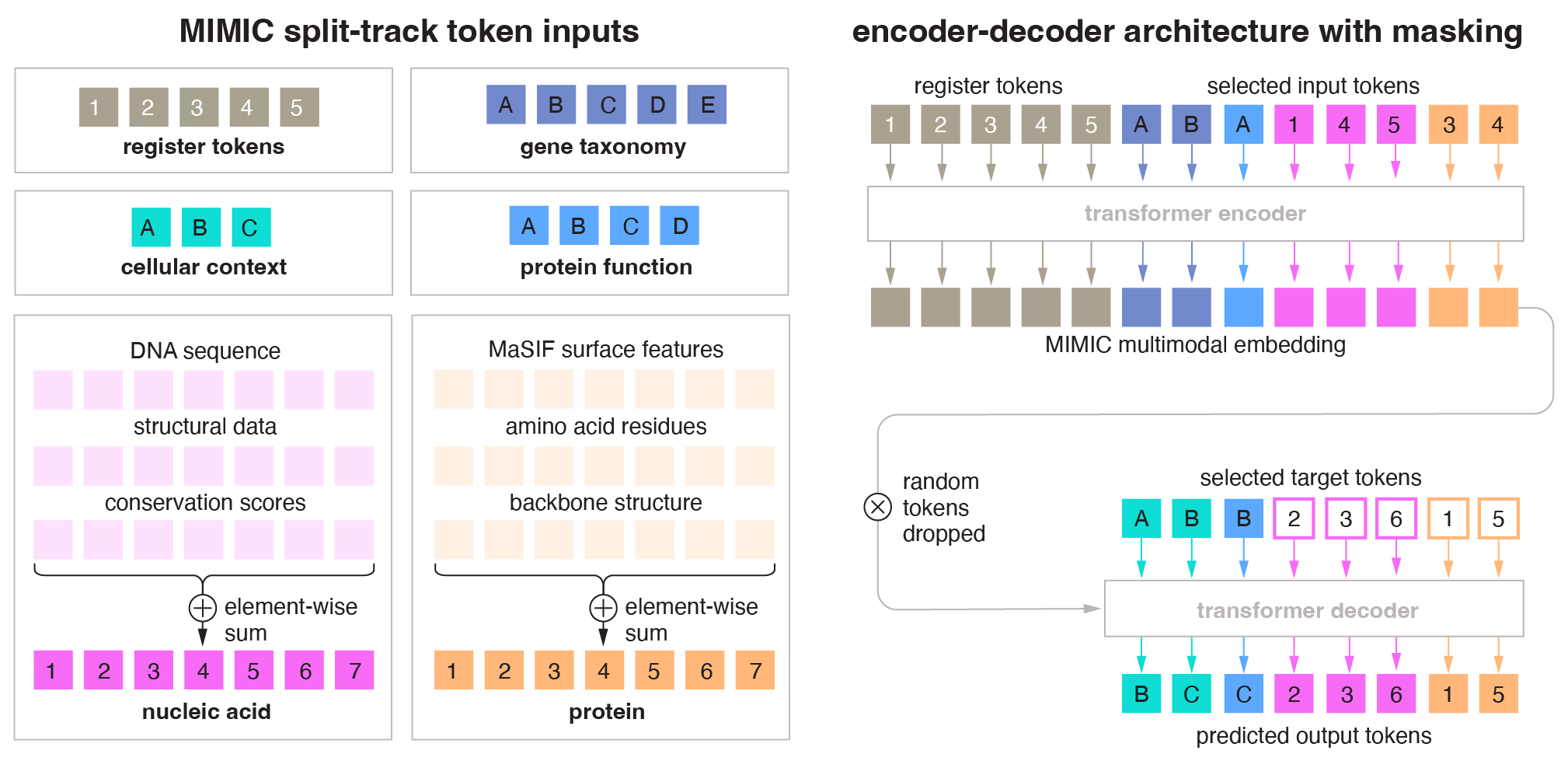}
    \caption{\textbf{The MIMIC Framework Architecture.} \textbf{(left)} Split-Track Input Construction: MIMIC manages multimodal heterogeneity by grouping biologically aligned features into ``Summed Tracks.'' The Nucleic Acid track element-wise sums sequence sequences ($L_{seq}$), structural data, and conservation scores into a single representation. Similarly, the Protein track utilizes the ESM3 tokenizer to sum amino acid residues with discretized 3D structure tokens, preserving vertical alignment. These biological tracks are concatenated with discrete semantic modalities (which occupy a Shared Semantic Embedding Space) and learnable Register Tokens to create the final unified token sequence. \textbf{(right)} Encoder-Decoder with Latent Dropout: Following a 4M-inspired backbone, the Transformer Encoder processes the unified input. A random token dropout is applied to the resulting latent state. The Transformer Decoder then probes this latent representation via cross-attention using target/query tokens to handle reconstruction objectives and generate global state summaries.}
    \label{fig:arch}
\end{figure}

\subsection{Unified encoder-decoder backbone via cross-attention}
MIMIC is built upon the 4M/AION encoder-decoder backbone \cite{4M,Parker2025AION1} with hyperparameterers  shown in \cref{tab:mimic_architecture_params}. All the experiments in this paper use the \texttt{x-transformers} configuration implemented in \cite{x-transformers} both for the encoder and decoder. MIMIC employs a full decoder that probes the encoder's latent representation via cross-attention. The self-attention blocks of both the decoder and the encoder have 50\% bidirectional and 25\% causal and 25\% anti-causal attention masks following the insight of~\cite{contcount} that causal attention can significantly outperform bidirectional attention in certain tasks.

\begin{table*}
\centering
\caption{\textbf{MIMIC Architecture Hyperparameters.} Summary of most relevant architectural choices.}
\begin{tabular}{p{0.34\linewidth}p{0.26\linewidth}}
\toprule
Architecture parameter & Value \\
\midrule
\midrule
Encoder depth & 20 layers \\
Decoder depth & 12 layers \\
Hidden width & 1536 \\
Attention heads & 24 per stack \\
Self Attention Mask & 50\% bidirectional, 25\% causal, 25\% anti-causal\\
Encoder token budget & 10,000 \\
Decoder token budget & 1,000 \\
Register tokens & 5 \\
RoPE fraction & 0.75 \\
\bottomrule
\end{tabular}
\label{tab:mimic_architecture_params}
\end{table*}

\subsection{Split-Track summation}\label{app:split-track}
A core innovation of MIMIC is the efficient packing of biological data. Rather than flattening all modalities into a single prohibitively long sequence, we group biologically aligned features into \textit{Summed Tracks}.

\begin{itemize}
    \item \textbf{Nucleic Acid Track:} We align the RNA/DNA sequence ($L_{seq}$) with its corresponding structural data and conservation scores. These are projected into the same latent dimension and summed element-wise.
    
    \item \textbf{Protein Track} We perform the same operation for the protein track with DSSP secondary structure and SASA naturally aligned with the protein sequence. For the backbone structure, we utilize the ESM3 structure tokenizer \cite{Hayes2025ESM3} to discretize 3D coordinates which are also aligned with the amino acid residues. Finally, we align MaSIF surface features~\cite{Gainza2020MaSIF} by assigning to each residue the sum of each surface feature's values associated to grid points that are closer than 2.8 angstrom to the residue's $C_\alpha$.
    
    \item \textbf{Discrete Modalities:} Semantic modalities such as cellular context descriptions or taxonimic information are treated as distinct, un-summed tracks. These are concatenated to the biological tracks, allowing the model to process ``metadata'' and ``molecular data'' as separate but interacting information streams.
\end{itemize}

\subsection{Register tokens and downstream representations}
\label{app:register}
To aggregate a global representation of the biological state, we append learnable \textbf{Register Tokens} to the sequence. These tokens serve as information ``sinks,'' pooling features from across the diverse tracks (DNA, Protein, Context) into a compact vector summary.

These register tokens also provide a convenient anchor for downstream representation learning. We recommend using register tokens together with a mean-pooled relevant track when a task requires a single vector representation, and using register tokens together with the full relevant track when the downstream task is token-level, such as residue-level or nucleotide-level prediction.

\subsection{Shared semantic embedding space}
To bridge the gap between hard experimental data and human-interpretable concepts, we unify all text-based modalities into a single shared embedding space. The context, functional captions, and taxonomy share weights with the biomedical corpus. This weight tying enables the model to transfer semantic reasoning learned from the large-scale corpus to the sparse context tokens.

\subsection{Localized Rotary Positional Embeddings (RoPE)}
We implement Rotary Positional Embeddings (RoPE) with a \textit{local group reset} strategy. The positional index is reset to 0 at the start of each distinct group. Crucially, this rotation is calculated relative to the element's position within its group, not its absolute index in the tensor. When tokens are dropped out during training, the remaining tokens retain their original positional indices, ensuring that the model preserves correct residue-to-residue distances regardless of input sparsity.

\section{Training}
\label{sec:training}

While the MIMIC architecture provides the tensor capability to process multimodal data, the core challenge lies in optimization: how to train a single model to simultaneously predict diverse biological tasks such as protein folding structure, genomic variant effect predictor, regulatory behavior. This is complicated by the extreme heterogeneity of biological data availability. A specific gene in the LORE dataset might have high-confidence AlphaFold structure and deep evolutionary alignments but lack cell-specific ATAC-seq data; another might have extensive transcriptomic profiling but no solved protein structure.

In total, we observe approximately 150 distinct ``modality presence signatures,'' or combinations of which tracks are available for a given sample, across the dataset. We employ a heuristic-based training strategy that simplifies these combinatorial possibilities into defined \textbf{Training Pathways}, optimized via an asymmetric token budget and a staged curriculum.

\subsection{Heuristic pathway sampling}
Rather than having a monolithic objective function, we define training as a stochastic sampling process over a set of $\approx 25$ heuristic pathways. Each pathway $P_k$ is a tuple defined as:
$$ P_k = (\mathcal{I}_{req}, \mathcal{I}_{opt}, \mathcal{T}_{req}, \mathcal{T}_{opt}, w_k) $$
where $\mathcal{I}$ denotes Input modalities, $\mathcal{T}$ denotes Target modalities (to be predicted), and $w_k$ is the sampling weight.

This heuristic approach allows us to group scientifically distinct tasks into simplified training routines. For example, to teach the model regulatory logic, we define pathways that mandate the presence of cellular context tags ($\mathcal{I}_{req} = \{\text{RNA, Context}\}$). However, to ensure the model remains robust to missing metadata, we also include ``context-naive'' pathways where the model must predict properties solely from sequence ($\mathcal{I}_{req} = \{\text{RNA}\}$). During a training step, a sample is drawn from the dataset. We then verify which pathways are valid for this sample's available modalities. A specific pathway is selected based on the normalized weights $w_k$, effectively up-sampling rare but high-value tasks (e.g., experimental structure prediction) relative to abundant tasks (e.g., simple sequence modeling).

\subsection{Asymmetric budgeting and target packing}
Biological modeling requires a massive receptive field to capture long-range interactions (e.g., enhancer-promoter loops), but the generation of dense structural data is computationally expensive. We address this with an asymmetric context window: 10,000 tokens for the Encoder and 1,000 tokens for the Decoder.

To maximize training efficiency within this 1,000 token decoder limit, we employ a \textbf{Target Packing} strategy. Many biological targets are sparse or short (e.g., splice junction tracks or single-value abundance measurements). Rather than training on these in isolation, we ``fill the budget'' by packing multiple disjoint targets into a single decoder pass until the 1,000-token limit is reached.

For instance, if the primary objective of a pathway is to predict Splice Junctions ($\approx 200$ tokens), and the sample also possesses evolutionary data, the data loader will automatically append the phyloP track ($\approx 800$ tokens) to the target sequence. This acts as a powerful regularizer, forcing the model to learn joint dependencies between splicing and conservation even when the explicit task was merely splicing prediction.

\subsection{Staged curriculum learning}
Training stability is critical when optimizing across such diverse modalities. We implemented a staged curriculum that progressively scales the context window from 1k $\to$ 2k $\to$ 4k $\to$ 8k $\to$ 10k tokens. This allows the model to learn local feature associations before attempting to resolve global dependencies. Crucially, the data handling strategy differs by modality during this scaling:

\begin{itemize}
    \item \textbf{RNA/DNA (Crop Strategy):} Nucleic acid sequences are randomly cropped to fit the current context window. Because RNA folding is often driven by local base-pairing, windowed crops remain valid training signals.
    \item \textbf{Proteins (Drop Strategy):} The fold of a protein and therefore its function is a holistic global property; a truncated protein sequence cannot reliably be used for structure prediction due to missing long-range contacts. Therefore, if a protein's structure track exceeds the current stage's token budget, the sample is dropped entirely rather than cropped.
\end{itemize}

Additionally, for the protein track, we explicitly align the coding sequence codons to the amino acid residues, treating \texttt{rna\_codons} as a distinct track.

\subsection{Dynamic workload balancing}
\label{sec:dynamic_workload}

\begin{figure}[ht!]
    \centering
    \includegraphics[width=\linewidth]{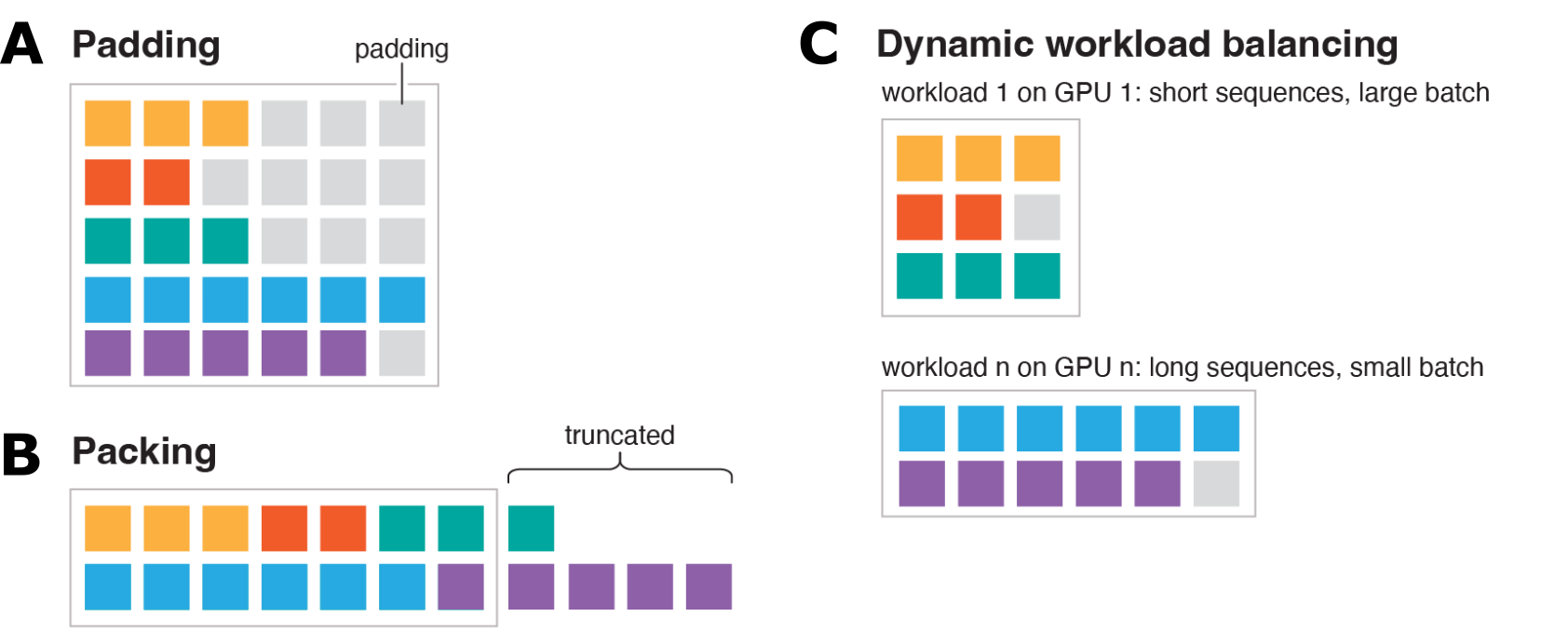}
    \caption{\textbf{Workload balancing strategies for multi-scale biological sequences.} 
    \textbf{(A)} Naive padding incurs significant computational waste (grey tokens) due to heavy-tailed distributions, where short peptides (cool colors) are padded to match long RNAs (warm colors). 
    \textbf{(B)} Sequence packing reduces padding but compromises biological integrity by truncating meaningful long-range context. 
    \textbf{(C)} Dynamic Workload Balancing (ours) buckets samples by length with adaptive batch sizes. This preserves single-entity semantics and coordinate frames without truncation while maximizing memory efficiency and I/O locality via bucket affinity.}
    \label{fig:workload_balancing}
\end{figure}

Our sequence lengths span multiple biological scales, from short peptides to long non-coding RNAs. With this heavy-tailed distribution, padding-based batching (padding each batch to its longest example, \cref{fig:workload_balancing}a) is extremely wasteful: a small fraction of long samples forces most batches to pay the memory and compute cost of the maximum length, dominated by padding rather than biological signal \cite{huggingface_data_collator}.

In large-scale NLP, this is often mitigated with sequence packing (example/sample packing, \cref{fig:workload_balancing}b): concatenating multiple shorter examples into fixed-length packs with boundary-aware masking to reduce padding.\cite{huggingface_sequence_packing,nemo_sequence_packing} In practice, packing pipelines frequently truncate or drop overflow.\cite{nemo_sequence_packing} This is incompatible with MIMIC: packing mixes multiple biological entities into one sequence, while our register tokens, targets, and multi-track alignment assume a single coherent entity and coordinate frame (e.g., CDS codons $\leftrightarrow$ residues; coordinate-tied regulatory tracks). Truncation is also not benign: it deletes meaningful long-range context (global folding contacts; distal regulation), creating systematic supervision corruption.

We therefore use length-bucketed dynamic batching aligned with the staged learning curriculum (\cref{fig:workload_balancing}c). At each stage, samples are bucketed by total token length (encoder+decoder), and batch size is chosen per bucket to balance peak memory against step time. Importantly, equal-token budgets do not imply equal compute: attention cost grows superlinearly with sequence length, so longer buckets are compute-heavier even at similar token counts. This emphasis differs from Dynamic Context Parallelism \cite{nvidia_dynamic_context_parallelism}, which addresses variable-length efficiency by adapting the degree of context parallelism at the systems level; our method is a data-side complement that reduces padding and preserves per-sample multimodal semantics.

To improve input throughput, we enforce \emph{bucket affinity}: each GPU (and thus each node) is pinned to a length bucket, repeatedly streaming from the same bucket-specific shard set. This improves I/O locality and cache reuse (page cache / file system cache and object-store read-ahead), reducing cross-bucket thrash and stabilizing step time.

\subsection{Register token self-supervision}
MIMIC utilizes learnable ``register tokens'' to aggregate global information. To prevent these tokens from degenerating into trivial copy mechanisms, we apply a \textbf{Random Token Dropout} of 0--10\% across all input tracks.

The register tokens are trained via a reconstruction objective: they must encode sufficient information to allow the decoder to recover the masked input tokens. This forces the registers to act as compressed ``state vectors'' for the biological entity, capturing high-level semantic and structural properties that are robust to local noise.

\subsection{Training hyperparameters}
Batching follows the length-bucketed dynamic batching strategy described in \cref{sec:dynamic_workload}, with per-bucket batch sizes chosen to balance memory usage and throughput. Training is distributed, by default, across 92 NVIDIA H200 GPUs using NCCL, although the number of GPUs used at each stage varied occasionally due to resources availability. Training proceeds for multiple epochs over the dataset with a staged curriculum that increases the maximum context length from 1k to 10k tokens. All training hyperparameters are summarized in \cref{tab:mimic_hyperparams_shared} for shared parameters between curriculum learning stages and in \cref{tab:mimic_hyperparams} for the individual stages.

\begin{table}[!ht]
\centering
\small
\setlength{\tabcolsep}{6pt}
\caption{\textbf{MIMIC training hyperparameters shared across stages.} Summary of optimization, batching, and distributed training settings used commonly across all stages of context-length curriculum training.}
\begin{tabular}{p{0.34\linewidth}p{0.15\linewidth}}
\toprule
Shared training hyperparameter & Value \\
\midrule
\midrule
Optimizer & AdamW \\
AdamW betas & $(0.9, 0.95)$ \\
AdamW epsilon & $1\times 10^{-8}$ \\
Weight decay & 0.05 \\
Learning-rate schedule & Cosine decay \\
Warmup / cooldown & 8 / 10 epochs \\
Numerical precision & \texttt{bfloat16} \\
\bottomrule
\end{tabular}
\label{tab:mimic_hyperparams_shared}
\end{table}

\begin{table}[!ht]
  \centering
  \small
  \caption{\textbf{MIMIC training hyperparameters by curriculum learning stages.} Summary of optimization, batching, and distributed training settings used across all stages of context-length curriculum training. The minimum effective learning rate is 100× smaller than maximum effective learning rate for all stages.}
  \begin{tabular}{p{0.28\linewidth}p{0.11\linewidth}p{0.11\linewidth}p{0.11\linewidth}p{0.11\linewidth}p{0.11\linewidth}}
  \toprule
  Training hyperparameter & 1k stage & 2k stage & 4k stage & 8k stage & 10k stage \\
  \midrule \midrule
  Effective max learning rate & $2.21\times 10^{-4}$ & $1.89\times 10^{-4}$ & $1.44\times 10^{-4}$ & $1.05\times 10^{-4}$ & $4.66\times 10^{-5}$ \\
  Total batch size & $\sim$ 2{,}000 & $\sim$ 2{,}300 & $\sim$ 2{,}600 & $\sim$ 3{,}000 & $\sim$ 3{,}100 \\
  \bottomrule
  \end{tabular}
  \label{tab:mimic_hyperparams}
  \end{table}

\section{Model evaluation}
\label{app:model_eval}

\subsection{Sequence completion using multimodal conditioning}
\label{app:seq-completion}

To perform the inpainting task, MIMIC is conditioned on the amino acid or nucleotide sequence, as well as additional auxiliary modalities for the length of a protein or unspliced transcript. A 100-token window in the center of the sequence is masked, and MIMIC is prompted to reconstruct this missing window. Sequences are reconstructed in a single model pass, by sampling from a softmax over the model's predicted logits (temperature $=1e-8$). All models are evaluated based on a per-token reconstruction accuracy; thus random performance for nucleotide reconstruction $\approx 0.25$, and for amino acid reconstruction $\approx 0.04$

Although multimodal sequence completion is a controlled benchmark rather than an biologically relevant goal, it provides a useful test of whether the model has learned the shared latent factors that constrain sequence under auxiliary observations, which informs decisions about MIMIC for both representation learning and design. Let \(A\) denote the masked nucleotide or amino-acid sequence and \(O\) a diverse set of auxiliary biophysical observations. When these observations substantially reduce uncertainty over sequence identity, the inpainting task is no longer a purely local denoising problem (i.e. \ when \(H(A \mid O) \ll H(A)\), predicting \(A\) under conditioning by \(O\) where $H(\cdot)$ is the Shannon entropy). Rather, it requires inferring the latent biochemical and structural factors that jointly constrain both sequence and observation. This view suggests that success on multimodal sequence completion should translate into more informative sequence representations for other downstream targets conditioned on sequence, provided those targets depend on overlapping latent factors even if they are not directly supervised during pretraining.


\begin{figure}[ht!]
    \centering
    \includegraphics[width=\linewidth]{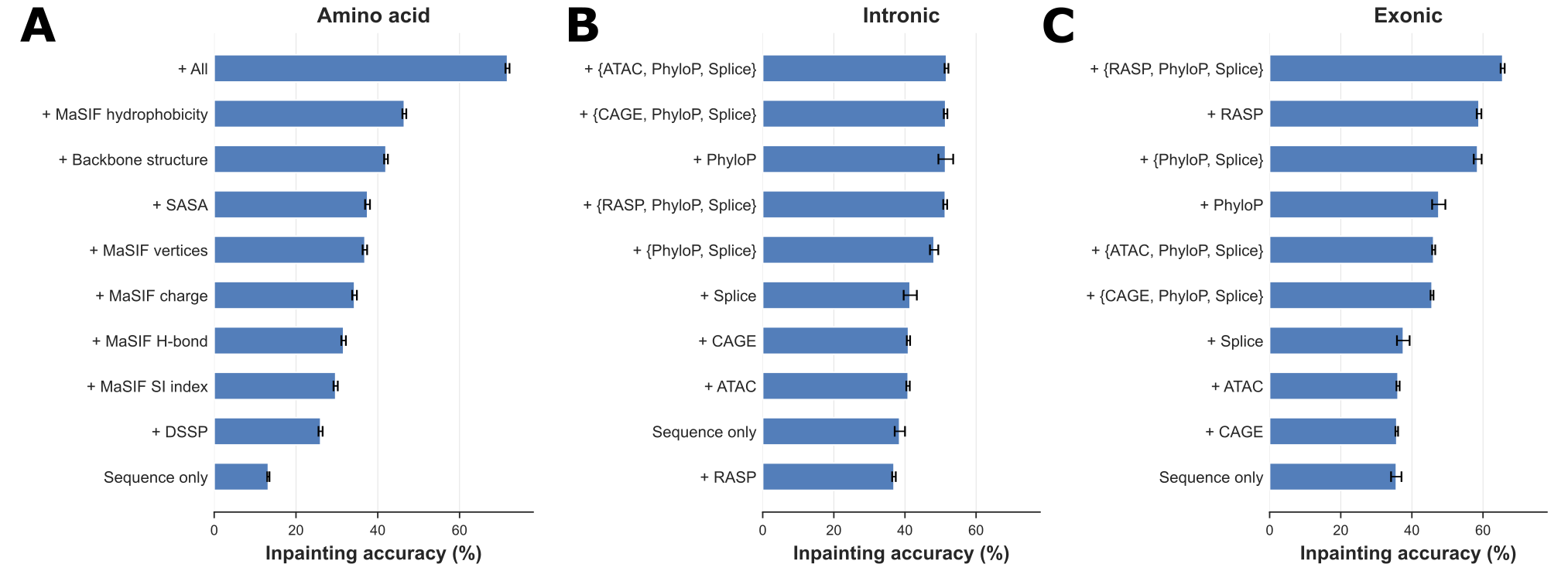}
    \caption{\textbf{Ablation on conditioning inputs to MIMIC.} Adding additional modalities as conditioning to MIMIC significantly increases its performance on the sequence completion task, demonstrating the value of a model that can ingest a multimodal context. \textbf{(C)} For amino acid sequences, MIMIC sees performance gains from each structure or chemical modality, and the largest increase in performance when all are provided as conditioning. \textbf{(B)} For intronic sequences, the strongest inpainting performance comes from including evolutionary conservation, splicing patterns, and epigenetic information. We note that RASP data is only available for exonic sequences, which is why performance does not meaningfully change. \textbf{(C)} For exonic sequences where mRNA 2D structure plays a meaningful role to constrain evolution, RASP data significantly improves inpainting performance.}
    \label{fig:inpainting_ablation}
\end{figure}

\paragraph{Ablation of modalities}
We explored the effect of ablating different input modalities in the sequence completion task to see which modalities contribute the most to model performance. For amino acid sequences (\cref{fig:inpainting_ablation}a), the most effective single modalities are the MaSIF hydrophobicity and tokenized backbone structure, with the combination of all structural and chemical modalities leading to a significant improvement over the addition of any single modality. For intronic sequences (\cref{fig:inpainting_ablation}b), the largest increase in performance comes from including evolutionary information (phyloP), splicing information, and epigenetics (either ATAC-seq or CAGE peaks). For exonic sequence (\cref{fig:inpainting_ablation}c), the largest increase from an individual modality is the RASP score, which provides structural information for mature mRNA molecules. Across all sequence types, while each individual modality provided a bump in performance, we saw continual gains from adding additional modalities, indicating that each provides complimentary information allowing the model to succeed at the inpainting task.

\paragraph{Exploring variable performance gain}
The improvement in performance from conditioning on an auxiliary modality is heterogeneous across different samples: in some cases, adding a modality yields substantial improvements, while in others the effect is negligible. To probe the source of this variability, we asked whether the multimodal conditioning performance increase is explained by how well the model can \textit{predict} the same modality from the nucleotide or amino acid sequence.

Concretely, for each masked segment used in the inpainting task, we additionally evaluate a matched \emph{modality-prediction} task on the same sequence interval. For a masked 100 amino acid (or nucleotide) span of the sequence, instead of conditioning on the sequence and auxiliary modality to reconstruct the masked segment, we provide the full sequence context and prompt the model to predict the modality in the masked interval. For protein sequence completion we use MaSIF hydrophibicity as the auxiliary modality, while for nucleic acid sequence we use RASP2; these two displayed the largest individual increase in performance in our previous ablation. We then compare MIMIC's modality-conditioned inpainting accuracy to its modality prediction accuracy.

\begin{figure}[ht!]
    \centering
    \includegraphics[width=\textwidth]{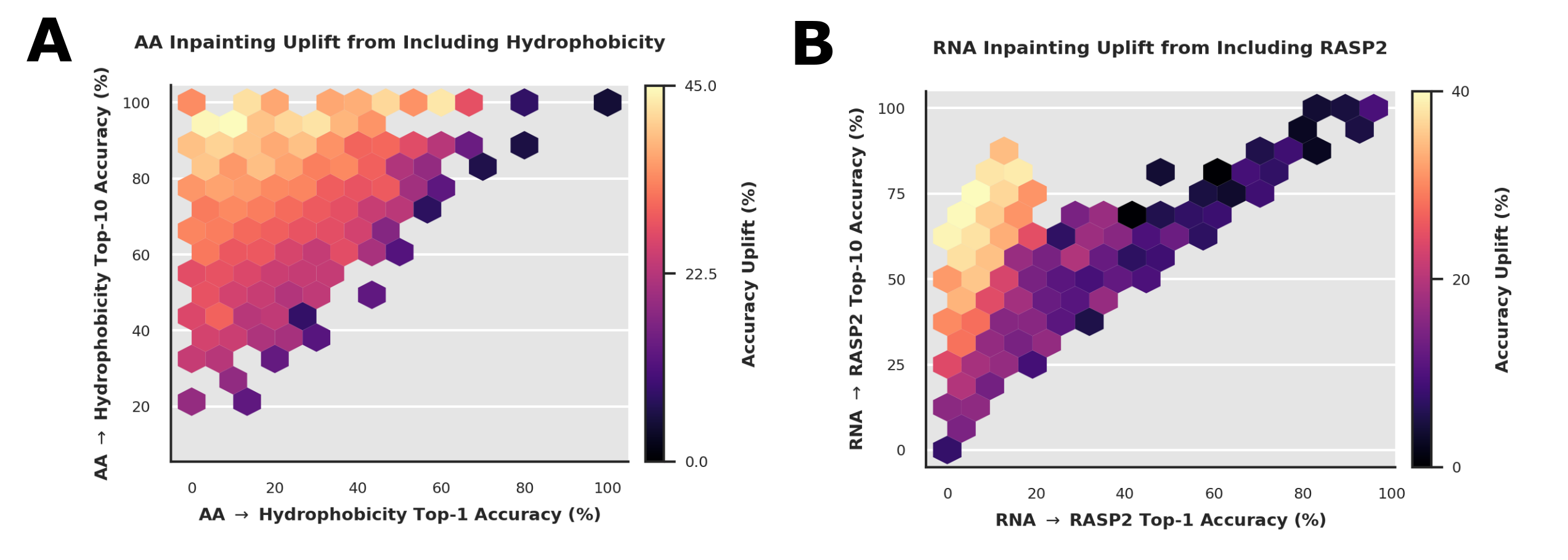}
    \caption{\textbf{Performance gains from multimodality depends on modality predictability.}
    \textbf{(A)} Each bin groups spans by AA$\rightarrow$hydrophobicity prediction accuracy (x: top-1; y: top-10) and is colored by AA inpainting accuracy increase (uplift) from conditioning on hydrophobicity. Uplift peaks when top-10 is high but top-1 is moderate, indicating that MaSIF hydrophobicity a provides complementary (non-redundant) constraint when it is only partially predictable from sequence alone.
    \textbf{(B)} We see the same trends for RNA inpainting. We show RNA$\rightarrow$RASP2 top-1 vs top-10 accuracy colored by accuracy uplift. Largest gains occur when top-10 is high but top-1 remains low, indicating complementary (non-redundant) constraint between sequence and RASP2.
    }
    \label{fig:uplift_attribution}
\end{figure}

We find that multimodal conditioning is most useful where the auxiliary modality is \textit{ambiguous} from sequence alone. That is, in regimes where the model has not learning anything about the sequence $\rightarrow$ modality relationship additional conditioning is unhelpful, but it is equally unhelpful when that relationship is trivial and little additional information is provided. We quantify this ``ambiguous'' regime by computing the top-1 and top-10 accuracy of MIMIC predicted bins for hydrophibicity (\cref{fig:uplift_attribution}a) and RASP2 (\cref{fig:uplift_attribution}b). The largest increase in inpainting accuracy appears for samples where top-10 accuracy is high while top-1 accuracy remains low. In this setting the model can narrow the auxiliary modality to a small set of plausible configurations but cannot resolve the exact state; providing the true measurement therefore removes residual ambiguity and yields the greatest improvement in reconstruction. Conversely, a high top-1 accuracy, either because the information is already present or the pattern is trivial, renders multimodal conditioning largely redundant and the performance increase diminishes.

\subsection{Protein embedding evaluation with PFMBench}
\label{app:pfmbench}

For each task in PFMBench \cite{pfmbench}, we trained a supervised probe on representations from MIMIC. Specifically, we prompted MIMIC with the amino acid sequence, and provided the full concatenation of the register tokens along with and protein-track outputs (preserving token-level information). We followed the training procedure outlined in the benchmark manuscript, including for hyperparameter selection. Performance metrics for competing models are taken from the benchmark manuscript. We report the 11 ``representative tasks'' sub-selected by the original authors. Full performance metrics for all tasks and models are reported in \cref{tab:pfmbench}.

\begin{sidewaystable}[ht!]
\centering
\caption{\textbf{Model performance on PFMBench.}  For each task, the best performing model is in \textbf{bold} and the second best is \textit{italic.} Metrics: Antibiotic resistance (Accuracy), Enzyme Commission (EC) number (F1 score), Localization (F1 score), Secondary structure (Accuracy), Stability (AUROC), BindingDB (Spearman), PDBbind (Spearman), Metal Ion Binding (Accuracy), Cloning CLF (AUROC), Material Production (Accuracy), Solubility (AUROC).}
\label{tab:pfmbench}
\begin{tabular}{lccccccccccc}
\toprule
 & \multicolumn{3}{c}{\textit{Function}} & \multicolumn{2}{c}{\textit{Structure}} & \multicolumn{3}{c}{\textit{Interaction}} & \multicolumn{3}{c}{\textit{Developability}} \\
\cmidrule(lr){2-4}\cmidrule(lr){5-6}\cmidrule(lr){7-9}\cmidrule(lr){10-12}
\textbf{Model} & \textbf{\thead{Antibiotic\\Resistance}} & \textbf{\thead{EC \#}} & \textbf{\thead{Localization}} & \textbf{\thead{Secondary\\Structure}} & \textbf{\thead{Stability}} & \textbf{\thead{BindingDB}} & \textbf{\thead{PDBbind}} & \textbf{\thead{Metal Ion\\Binding}} & \textbf{\thead{Clone\\CLF}} & \textbf{\thead{Material\\Production}} & \textbf{\thead{Solubility}} \\
\midrule
\midrule
MIMIC & \textit{0.688} & 0.595 & 0.746 & 0.808 & 0.307 & \textbf{0.267} & \textbf{0.204} & 0.736 & \textbf{0.849} & \textbf{0.835} & \textbf{0.857} \\
\midrule
DPLM \cite{wang2024diffusion}    & 0.687          & 0.755          & 0.758          & 0.757          & 0.294          & 0.174          & 0.137          & 0.701          & 0.812          & 0.801          & 0.828 \\
ESM2 \cite{esm2}   & 0.634          & 0.736          & \textit{0.762} & 0.764          & 0.321          & 0.137          & 0.147          & 0.712          & 0.806          & 0.812          & \textit{0.845} \\
ESM3 \cite{Hayes2025ESM3}    & 0.584          & 0.648          & 0.659          & 0.813          & 0.157          & \textit{0.225} & 0.156          & 0.703          & 0.774          & 0.775          & 0.781 \\
ESM-C \cite{esm-c}   & 0.673          & 0.717          & 0.754          & 0.768          & 0.300          & 0.207          & 0.147          & 0.702          & 0.810          & 0.810          & 0.842 \\
PGLM \cite{xtrimopglm}     & 0.673          & 0.747          & 0.748          & 0.728          & \textit{0.331} & 0.169          & 0.169          & 0.745          & \textit{0.836} & 0.795          & 0.822 \\
ProtGPT2 \cite{ferruz2022protgpt2} & 0.684          & 0.697          & 0.703          & 0.494          & 0.148          & 0.172          & 0.135          & 0.712          & 0.777          & 0.768          & 0.789 \\
ProstT5 \cite{heinzinger2023prostt5}  & \textbf{0.690} & \textbf{0.768} & 0.732          & \textit{0.814} & 0.130          & 0.166          & 0.183          & 0.721          & 0.799          & 0.816          & 0.819 \\
ProtST \cite{xu2023protst}  & 0.634          & 0.718          & 0.749          & 0.685          & 0.066          & 0.189          & 0.195          & 0.515          & 0.807          & 0.693          & 0.820 \\
ProTrek \cite{su2025trimodal} & 0.593          & \textit{0.764} & \textbf{0.839} & 0.774          & 0.049          & 0.192          & 0.173          & \textbf{0.804} & 0.826          & 0.815          & 0.834 \\
ProtT5 \cite{elnaggar2021prottrans}   & 0.687          & 0.762          & 0.726          & 0.780          & 0.186          & 0.197          & \textit{0.201} & 0.721          & 0.785          & 0.801          & 0.787 \\
SaProt \cite{su2023saprot}  & 0.658          & 0.751          & 0.740          & \textbf{0.824} & 0.248          & 0.166          & 0.155          & \textit{0.763} & 0.812          & 0.811          & 0.844 \\
VenusPLM \cite{tan2025venusfactory} & 0.646          & 0.752          & 0.738          & 0.716          & \textbf{0.339} & 0.168          & 0.165          & 0.702          & 0.832          & \textit{0.820} & 0.828 \\
\bottomrule
\end{tabular}
\end{sidewaystable}

\subsection{RNA embedding evaluation with mRNABench}
\label{app:mrnabench}

Likewise, for mRNABench \cite{mrnabench} we trained a supervised probe on MIMIC representations. Because mRNABench does not have any nucleotide-level tasks, we trained the probe on the concatenation of the register tokens and a mean-pooled RNA-track representation. MIMIC was trained on unspliced transcripts, while mRNABench contains mature (spliced) mRNA sequences and, where relevant, a corresponding coding sequence track. To address this difference, we leverage the multimodal nature of MIMIC to generate representations under several conditioning cases. In the non-coding case, we generate embeddings using only nucleotide (NT) sequence. In the coding case, we generate embeddings using (1) NT sequence only, (2) NT and protein (AA) sequence (3) NT sequence, codons, and AA sequence.

For each case, we then augment the mRNABench training data with additional samples containing each subset of input modalities (e.g. NT, AA, and NT+AA for case 2 above). At inference time, each subset is used to generate embeddings per sample, and the average probe prediction across all subsets of modalities is used as the model prediction for that sample. The mRNABench validation set is used to select the best inference case per task. Performance metrics for competing models are taken from prior work and are not recomputed here. Full performance metrics for all tasks and models are reported in \cref{tab:mrna_bench}.

\begin{table}[ht!]
\centering
\caption{\textbf{Model performance on mRNABench.} For each task, the best performing model is in \textbf{bold} and the second best is \textit{italic.} Metrics: eCLIP (AUPR), GO (AUPR), Protein Localization (AUPR), RNA Localization (AUPR), MRL LBKWK (Pearson), MRL Sugimoto (Pearson), RNA Half Life (Pearson).}
\label{tab:mrna_bench}
\begin{tabular}{lccccccc}
\toprule
\textbf{} & \multicolumn{2}{c}{\textit{Function}} & \multicolumn{2}{c}{\textit{Localization}} & \multicolumn{3}{c}{\textit{Translation Regulation}} \\
\cmidrule(lr){2-3}\cmidrule(lr){4-5}\cmidrule(lr){6-8}
\textbf{Model} & \textbf{\thead{eCLIP}} & \textbf{\thead{GO}} & \textbf{\thead{Protein\\Localization}} & \textbf{\thead{RNA\\Localization}} & \textbf{\thead{MRL\\(LBKWK)}} & \textbf{\thead{MRL\\(Sugimoto)}} & \textbf{\thead{RNA\\Half Life}} \\
\midrule
\midrule
MIMIC      & \textit{0.470} & \textbf{0.475} & \textbf{0.415} & 0.756          & \textit{0.585} & \textbf{0.496} & 0.623 \\
\midrule
AIDO-RNA \cite{zou2024large}      & 0.420          & 0.366          & 0.343          & 0.734          & 0.534          & 0.360          & 0.555 \\
Dilated ResNet \cite{mrnabench} & 0.421          & 0.155          & 0.169          & 0.645          & 0.411          & 0.448          & 0.647 \\
DNABERT2 \cite{Zhou2023DNABERT2}      & 0.371          & 0.326          & 0.325          & 0.675          & 0.534          & 0.286          & 0.475 \\
DNABERT-S \cite{zhou2025dnabert}     & 0.374          & 0.318          & 0.311          & 0.685          & 0.393          & 0.292          & 0.467 \\
ERNIE-RNA \cite{yin2025ernie}     & 0.373          & 0.341          & 0.338          & 0.691          & 0.503          & 0.328          & 0.518 \\
Evo2 \cite{Brixi2025Evo2}          & \textbf{0.473} & \textit{0.467} & \textit{0.408} & \textit{0.767} & 0.507          & 0.455          & \textit{0.661} \\
HyenaDNA \cite{Nguyen2023HyenaDNA}      & 0.374          & 0.308          & 0.311          & 0.670          & 0.506          & 0.287          & 0.462 \\
NT-v2  \cite{dalla2025nucleotide}        & 0.391          & 0.350          & 0.336          & 0.719          & 0.512          & 0.297          & 0.530 \\
Orthrus \cite{fradkin2025orthrus}      & 0.465          & 0.435          & 0.396          & \textbf{0.789} & \textbf{0.633} & \textit{0.460} & \textbf{0.696} \\
RiNALMo \cite{penic2025rinalmo}       & 0.393          & 0.347          & 0.327          & 0.690          & 0.466          & 0.416          & 0.532 \\
SpliceBERT \cite{chen2024self}    & 0.377          & 0.382          & 0.360          & 0.726          & 0.514          & 0.336          & 0.531 \\
\bottomrule
\end{tabular}
\end{table}

\subsection{Model performance on variant effect prediction benchmarks}
\label{app:vep}

Because MIMIC is given only a nucleotide sequence and its local multimodal context, its phyloP output should not be interpreted as a reconstruction of the underlying phylogenetic calculation. This predicted quantity is better interpreted as a sequence-derived estimate of \textit{expected} evolutionary constraint: a learned mapping from local sequence context to the conservation level with which that context is most often associated in the training data. Under this interpretation, predicted values near zero correspond to sequence neighborhoods that resemble weakly constrained sites, positive values correspond to neighborhoods characteristic of constrained positions, and negative values correspond to neighborhoods associated with faster-than-neutral or weakly maintained sequence classes \cite{Pollard2010PhyloP}.


\paragraph{Using predicted phyloP for variant effects}
\label{subsec:constraint_disruption}

Given this constraint signal, we hypothesized that MIMIC-predicted phyloP would be useful to evaluate the functional impact of genomic sequence variants. To construct an informative multimodal signal of this impact, we predict not only the phyloP for the variant sequence, but the \textit{change} induced by mutation. Let $\widehat{p}^{\mathrm{wt}}_i$ and $\widehat{p}^{\mathrm{mut}}_i$ denote the model-predicted phyloP values at position $i$ for the wild-type and mutant sequence, respectively, and define the mutation-response profile
\[
\Delta \widehat{p}_i = \widehat{p}^{\mathrm{mut}}_i - \widehat{p}^{\mathrm{wt}}_i.
\]
For scalar variant scoring, we summarize this local response by the mean absolute perturbation in a centered window $W=30$ nucleotides around the variant,
\[
s_{\Delta p}(W) = \frac{1}{|W|} \sum_{i \in W} \left| \Delta \widehat{p}_i \right|.
\]


This variant-level formulation is important because raw conservation is not synonymous with pathogenicity. A synonymous substitution at a conserved position need not be pathogenic if it does not materially perturb splicing or other regulatory logic, whereas pathogenic splice-altering variants can arise in intronic sequences whose background conservation is only modest \cite{Zeng2021SynVep,VazDrago2017DeepIntronic,Kurosawa2023PDIVAS}. A mutation-dependent, context-aware quantity is therefore more directly aligned with variant effect than the reference-site score alone.

We refer to this score throughout the manuscript as ``MIMIC (phyloP VEP),'' which is used for prediction of wild-type alleles in ClinVar variants (\cref{fig:vep_supplement}), for the VEP benchmarks of mRNABench (\cref{fig:validation}f, \cref{tab:mrna_bench_vep}), and to evaluate whether MIMIC detects a splice-altering pathogenic variant (\cref{fig:splicing}e, middle). 

\paragraph{Pathogenicity signal in conserved and weakly conserved sequence}
\label{subsec:constraint_pathogenicity}

\begin{figure}[ht!]
    \centering
    \includegraphics[width=\linewidth]{figures/FigureVariant.png}
    \caption{\textbf{MIMIC predicted phyloP variant effect prediction provides complementary signal to raw conservation.} \textbf{(A)} Wild-type allele exact recovery rate from ClinVar pathogenic and benign SNVs, comparing sequence-only and sequence+phyloP conditioning; phyloP conditioning yields significant gains for both variant classes. \textbf{(B)} Normalized AUPRC for pathogenicity prediction stratified by the sign of observed phyloP. Raw phyloP is most informative at positively constrained positions (phyloP $\geq$ 0), while the MIMIC phyloP VEP score retains significant predictive value at weakly conserved or accelerated positions (phyloP $<$ 0), demonstrating complementarity between conservation and mutation-induced perturbation signals.}
    \label{fig:vep_supplement}
\end{figure}

To illustrate the value of multimodal conditioning and the MIMIC phyloP VEP score, we performed a study of variants in ClinVar. For each pathogenic location we mask the allele at this location and prompt MIMIC to predict the wild type allele, either from the local context only, or from the sequence context and phyloP score. Providing evolutionary constraint to MIMIC significantly increases the rate of correctly recovering the wild type allele (\cref{fig:vep_supplement}a). This improvement is even more pronounced for pathogenic variants, suggesting that constraint is an especially strong signal in these cases.

Thus, we next compared the wild type phyloP constraint with the MIMIC phyloP VEP predictor as a signal of variant pathogenicity, measured using normalized AUPR. We stratify these results by positions under positive constraint (phyloP $\geq 0$) versus those under weak constraint or accelerated evolution (phyloP $< 0$) (\cref{fig:vep_supplement}b). In the former regime, true phyloP outperforms MIMIC phyloP VEP, confirming that positive constraint is a strong pathogenicity signal. However, in the latter true phyloP loses nearly all its discriminative power, while MIMIC phyloP VEP is still able to accurately predict pathogenicity (although less so than in the highly-constrained regime).

This pattern indicates complementarity rather than redundancy. Reference-site conservation is most informative in regions already under detectable evolutionary constraint, as expected from comparative genomics \cite{LindbladToh2011, Pollard2010Rates}. In contrast, the mutation-induced response contributes most in sequence where background conservation is weak and pathogenicity depends more directly on mutation-specific disruption of local regulatory logic. This is precisely the regime in which splice-altering and other non-coding regulatory variants are often difficult to prioritize from conservation alone, despite the broad importance of non-coding disease-associated variation \cite{Jaganathan2019SpliceAI,Zeng2022Pangolin,Cormier2022ConSpliceML,VazDrago2017DeepIntronic,Kurosawa2023PDIVAS,farh2015noncoding}. The result therefore supports a broader interpretation of the phyloP head: it does not simply recover the observed conservation track, but learns sequence features whose perturbation is informative for functional variant effect.


This mutation-aware interpretation also provides a rationale for phyloP conditioning during generation (\cref{sec:corrective_design_splice}). If pathogenic variants are associated with departure from the local constraint regime expected for the wild-type sequence, then conditioning design toward a wildtype-like, non-pathogenic phyloP profile should bias generation away from deleterious regulatory configurations. 


\begin{table}[ht!]
\centering
\caption{\textbf{Variant effect prediction tasks from mRNABench.}  For each task, the best performing model is in \textbf{bold} and the second best is \textit{italic.} All metrics reported are AUPR.}
\label{tab:mrna_bench_vep}
\begin{tabular}{lcc}
\toprule
\textbf{Model} & \textbf{\thead{Complex Variants}} & \textbf{\thead{Mendelian Variants}} \\
\midrule
\midrule
MIMIC          & \textit{0.142} & 0.263 \\
MIMIC (phyloP VEP) & \textbf{0.253} & 0.500 \\
\midrule
AIDO-RNA \cite{zou2024large}      & 0.092 & 0.540 \\
Dilated ResNet & 0.080 & \textbf{0.585} \\
ERNIE-RNA \cite{yin2025ernie}     & 0.097 & 0.481 \\
Evo2 \cite{Brixi2025Evo2}         & 0.081 & 0.561 \\
DNABERT2 \cite{Zhou2023DNABERT2}      & 0.093 & 0.435 \\
DNABERT-S \cite{zhou2025dnabert}     & 0.070 & 0.542 \\
HyenaDNA \cite{Nguyen2023HyenaDNA}      & 0.064 & 0.320 \\
NT-v2  \cite{dalla2025nucleotide}        & 0.064 & 0.540 \\
Orthrus \cite{fradkin2025orthrus}       & 0.071 & \textit{0.574} \\
RiNALMo \cite{penic2025rinalmo}       & 0.086 & 0.546 \\
SpliceBERT \cite{chen2024self}    & 0.104 & 0.460 \\
\bottomrule
\end{tabular}
\end{table}

\subsection{Splice site prediction}

We evaluate MIMIC's ability to predict splice junctions and compare to SpliceAI \cite{Jaganathan2019SpliceAI}, NTv3 \cite{Boshar2025NucleotideTransformerV3} and AlphaGenome \cite{Avsec2026AlphaGenome}. To enable a fair comparison despite differing native input lengths, each model is supplied with the context required by its architecture: SpliceAI receives 5kb of flanking sequence on each side of the evaluation region, AlphaGenome is queried on a 16,384 bp genomic interval containing the evaluation window, NTv3 is queried on a genomic window sized to $\left\lceil \text{pred\_len}/0.375 \right\rceil$ (rounded up to the next multiple of 128 bp) and centered on the target region, and MIMIC consumes the full sequence up to 10kb directly.

MIMIC's splice junction head outputs a 5-class distribution over {non-splice, donor, acceptor, TSS, TES} at each position. Since the prediction region excludes TSS and TES anchor positions by construction, we compute per-position donor and acceptor probabilities via a 2-class softmax restricted to {non-splice, donor} and {non-splice, acceptor} logits respectively.

For both tasks described below, we evaluate model performance using Area Under the Precision and Recall Curve (AUPR). Since this task exhibits extreme class imbalance ($\sim$99.85\% of all nucleotide positions are not splice sites), we require a metric that can measure how well the models identify rare events. Additionally, a threshold-independent ranking metric like AUPR is appropriate because we are interested not only in whether the model assigns higher scores to true splice sites than to non-splice sites, but also in how well those scores support discrimination across decision thresholds. 

Since all models are trained with different held out portions, finding an overlapping test set greatly reduces the number of samples available. Therefore, all results presented here are on MIMIC's test set, but not necessarily the test sets of SpliceAI, NTv3, and Alphagenome. 

\paragraph{Gene-level Splice Prediction}
For the gene-level splicing task, we assemble a dataset of human coding and non-coding transcripts, and collect all possible splice junctions from every transcript belonging to a gene. For each gene, the evaluation region spans from the earliest transcription start site (TSS) to the latest transcription end site (TES) across all of its transcripts, and every position within this span is labeled as a donor, an acceptor, or neither based on the union of junctions observed in any transcript. Because many genes exceed MIMIC's context length, we restrict this exploration to genes with length < 25kb, and for genes longer than 10kb we evaluate all models on a common 10kb region with the most splice junctions. Samples are stratified by gene length to ensure balanced coverage across length bins, and stratified bootstrapping is used to construct error bars. Results are shown in \cref{fig:splicing}a above and \cref{tab:splice_benchmark} below.

\paragraph{Transcript-level Conditional Splice Prediction}
For the transcript-level task, we group transcripts that share the same TSS and TES into a single \emph{mimic-gene}, treating each mimic-gene as one evaluation unit. Donor and acceptor labels are defined as the union of junctions across all transcripts in the group, restricted to the interval between the shared TSS and TES. We first evaluate all models on this new dataset in the same unconditioned mode as in the gene-level task. Then, MIMIC is explicitly conditioned on the TSS and TES positions: the boundary tokens marking the start and end of the transcript are provided as input, along with 200 bp of flanking genomic context on each side.
As before, samples are stratified by transcript length to balance coverage across length bins and stratified bootstrapping provides error bars. Detailed results are given in \cref{fig:splicing}b3 above and \cref{tab:splice_benchmark} below.

\begin{table}[ht!]
\caption{\textbf{Model performance on human splice-site prediction.}  For each task, the best performing model is in \textbf{bold} and the second best is \textit{italic.} All metrics reported are AUPR.}
\label{tab:splice_benchmark}
\begin{tabular}{lcccc}
\toprule
 & \multicolumn{2}{c}{\textbf{Gene-level}} & \multicolumn{2}{c}{\textbf{Transcript-level}} \\
\cmidrule(lr){2-3}\cmidrule(lr){4-5}
\textbf{Model} & \textit{Coding} & \textit{Non-coding} & \textit{Coding} & \textit{Non-coding} \\
\midrule
\midrule
MIMIC (unconditioned) & \textbf{0.905 $\pm$ 0.009} & \textbf{0.641 $\pm$ 0.015} & \textit{0.847 $\pm$ 0.009} & \textit{0.669 $\pm$ 0.011} \\
MIMIC (conditioned) & -- & -- & \textbf{0.874 $\pm$ 0.008} & \textbf{0.701 $\pm$ 0.011} \\
AlphaGenome & \textit{0.898 $\pm$ 0.009} & \textit{0.572 $\pm$ 0.014} & 0.803 $\pm$ 0.009 & 0.531 $\pm$ 0.015 \\
NTv3        & 0.818 $\pm$ 0.015 & 0.156 $\pm$ 0.014 & 0.766 $\pm$ 0.011 & 0.272 $\pm$ 0.018 \\
SpliceAI    & 0.879 $\pm$ 0.014 & 0.562 $\pm$ 0.014 & 0.804 $\pm$ 0.009 & 0.577 $\pm$ 0.012 \\
\bottomrule
\end{tabular}
\end{table}

\subsection{RASP2 and context evaluation}
\begin{figure}[!ht]
    \centering
    \includegraphics[width=0.7\linewidth]{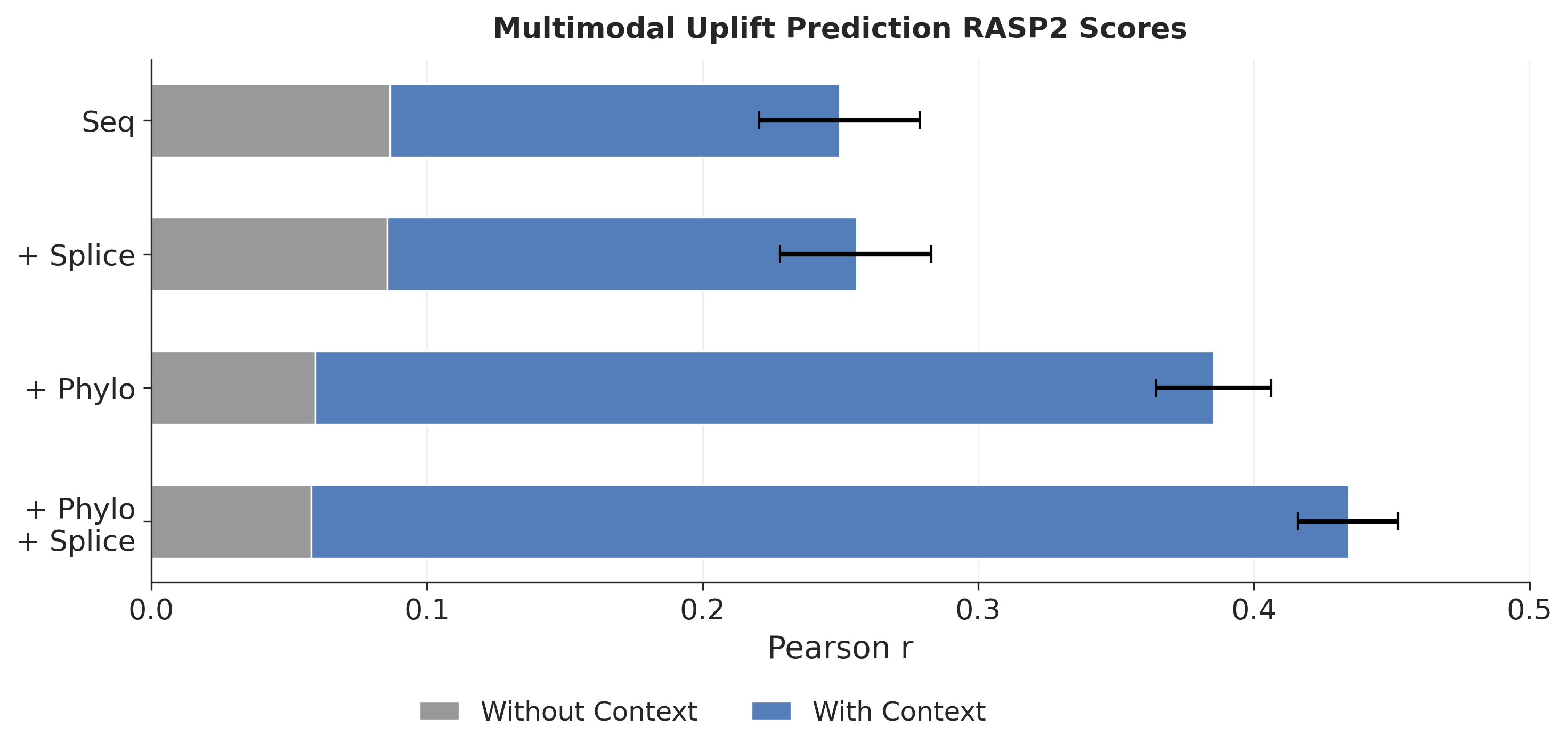}
    \caption{\textbf{MIMIC-predicted RASP2 scores under partial sequence observation and multimodal conditioning}. Predictions are conditioned on a contiguous window covering 50\% of the input RNA sequence, together with auxiliary modalities. Conditioning on phyloP and splicing, in addition to sequence context, yields the largest performance improvement over sequence-only baselines. All sequences are filtered to length $< 10$k nucleotides.}
    \label{fig:phylop_rasp_supp}
\end{figure}

We further analyze the contribution of contextual and multimodal conditioning to RASP2 score prediction in a setting where sequence information is partially observed and complemented by auxiliary tracks, including phyloP and splicing patterns (\cref{fig:phylop_rasp_supp}).
In contrast to the analysis in \cref{fig:rasp_context}, we evaluate on a modified dataset in which each input RNA sequence is restricted to a contiguous window covering 50\% of its full length. This controlled reduction in sequence information enables a more direct assessment of how non-sequence modalities contribute to prediction performance. In this regime, the model must rely on phyloP and splicing signals to compensate for missing sequence context when inferring RASP2 scores. This setup amplifies the relative contribution of auxiliary modalities compared to the full-sequence setting, where sequence information alone can dominate prediction. We observe that conditioning on phyloP and splicing jointly with the available sequence context yields the largest improvement in RASP2 prediction accuracy, highlighting the complementary nature of evolutionary and regulatory signals in constraining RNA structure.

\subsection{SHAPE-Guided RNA secondary structure modeling and evaluation}
\label{app:shape_folding}

To evaluate the biological utility of our predicted per-nucleotide reactivity scores, we assessed whether they can serve as effective surrogate experimental data for guiding thermodynamic RNA secondary structure prediction. We utilized \texttt{RNAfold} from the ViennaRNA package \cite{lorenz2011viennarna} incorporating SHAPE (Selective 2'-hydroxyl acylation analyzed by primer extension) reactivity constraints, comparing the resulting predicted structures against structures generated using experimental measured reactivities. Crucially, in this evaluation, the reference is defined as the minimum free energy (MFE) secondary structure computed by \texttt{RNAfold} when constrained by actual experimental chemical probing data, rather than experimentally derived 3D structures.

\paragraph{Evaluation Dataset} We randomly sampled 1,000 RNA transcripts from a held-out validation set. For a transcript of length $L$, each sample consists of the raw RNA sequence; ground-truth per-nucleotide reactivity values, denoted $\mathbf{r}^{\text{gt}} \in \mathbb{R}^L$, derived from experimental measurements (e.g., icSHAPE or related chemical probing technologies); and predicted per-nucleotide reactivity values, denoted $\hat{\mathbf{r}} \in \mathbb{R}^L$, output by MIMIC for the same transcript conditioned on the corresponding experimental context. Prior to folding, both the ground-truth and predicted reactivity scores are strictly normalized to a scale of $0.0$ to $1.0$.

\paragraph{SHAPE-Guided Folding Procedure} For each of the 1,000 transcripts, three independent secondary structure predictions were performed using \texttt{RNAfold} (version 2.7.2):
\begin{enumerate}
    \item \textbf{Ground-truth SHAPE run (\texttt{run\_gt}):} \texttt{RNAfold} was invoked with the ground-truth reactivity file via the \texttt{--shape} flag. The reactivity file was formatted as a two-column tab-delimited text file containing 1-based nucleotide indices and reactivity values.
    \item \textbf{Predicted SHAPE run (\texttt{run\_pred}):} \texttt{RNAfold} was invoked identically, but the \texttt{--shape} flag pointed to MIMIC's predicted reactivity values ($\hat{\mathbf{r}}$) in the same format.
    \item \textbf{Sequence-only baseline (\texttt{run\_seq\_alone}):} \texttt{RNAfold} was invoked with no SHAPE data, folding purely on the basis of thermodynamic nearest-neighbor free energies. This serves as the baseline against which SHAPE-guided improvements are measured.
\end{enumerate}

The \texttt{--shape} flag instructs \texttt{RNAfold} to incorporate per-nucleotide reactivities as pseudo-energy penalties using the model proposed by Deigan \textit{et al.} \cite{deigan2009accurate}. This model adds a pseudo-energy term $\Delta G_{\text{SHAPE}}(i)$ to the folding free energy of each paired nucleotide $i$:
\begin{equation}
    \Delta G_{\text{SHAPE}}(i) = m \cdot \ln(r_i + 1) + b
\end{equation}
where $r_i$ is the reactivity, and $m$ and $b$ are empirically determined slope and intercept parameters (using ViennaRNA default values). The minimum free energy (MFE) secondary structure is extracted from each \texttt{RNAfold} output as the dot-bracket string.

\paragraph{Base Pair Extraction and Evaluation Metrics} Secondary structures in dot-bracket notation are parsed into 0-based base pair sets $\mathcal{P} \subset \{(i,j) \mid i < j\}$ using a standard stack-based algorithm. 

We evaluated the predicted ($\hat{\mathcal{P}}$ from \texttt{run\_pred}) and sequence-only ($\mathcal{P}^{\text{seq}}$ from \texttt{run\_seq\_alone}) base pair sets against the ground-truth reference ($\mathcal{P}^{\text{gt}}$ from \texttt{run\_gt}). For a given prediction $\mathcal{P}$, True Positives (TP), False Positives (FP), and False Negatives (FN) are defined as:
\begin{equation}
    \text{TP} = |\mathcal{P} \cap \mathcal{P}^{\text{gt}}|, \quad \text{FP} = |\mathcal{P} \setminus \mathcal{P}^{\text{gt}}|, \quad \text{FN} = |\mathcal{P}^{\text{gt}} \setminus \mathcal{P}|
\end{equation}

Performance is quantified using Positive Predictive Value (PPV) and Sensitivity:
\begin{equation}
    \text{PPV} = \frac{\text{TP}}{\text{TP} + \text{FP}}, \qquad \text{Sensitivity} = \frac{\text{TP}}{\text{TP} + \text{FN}}
\end{equation}
yielding the harmonic mean, the $F_1$ score (where PPV, Sensitivity, and $F_1$ default to $0$ if denominators are $0$):
\begin{equation}
    F_1 = \frac{2 \cdot \text{PPV} \cdot \text{Sensitivity}}{\text{PPV} + \text{Sensitivity}}
\end{equation}

The structural improvement provided by MIMIC is measured as the difference in $F_1$ scores:
\begin{equation}
    \Delta F_1 = F_1(\hat{\mathcal{P}}, \mathcal{P}^{\text{gt}}) - F_1(\mathcal{P}^{\text{seq}}, \mathcal{P}^{\text{gt}})
\end{equation}
where $\Delta F_1 > 0$ indicates that incorporating predicted reactivities yields a structure closer to the experimental reference than sequence-alone folding.

\paragraph{Statistical Testing} To assess whether the improvement is statistically significant at the population level, we applied the Wilcoxon signed-rank test to the paired F1 scores $\left(F_1^{\text{pred}}, F_1^{\text{seq}}\right)_{i=1}^{1000}$ across all 1,000 transcripts. We report a one-sided $p$-value testing the alternative hypothesis that $F_1^{\text{pred}} > F_1^{\text{seq}}$. This was obtained by halving the two-sided $p$-value (applicable given that the observed median of $\Delta F_1$ is positive).

\section{Multimodal generative design}

\subsection{RNA splicing design}
\label{app:rna_splice_design}

\begin{figure}[ht!]
    \centering
    \includegraphics[width=\linewidth]{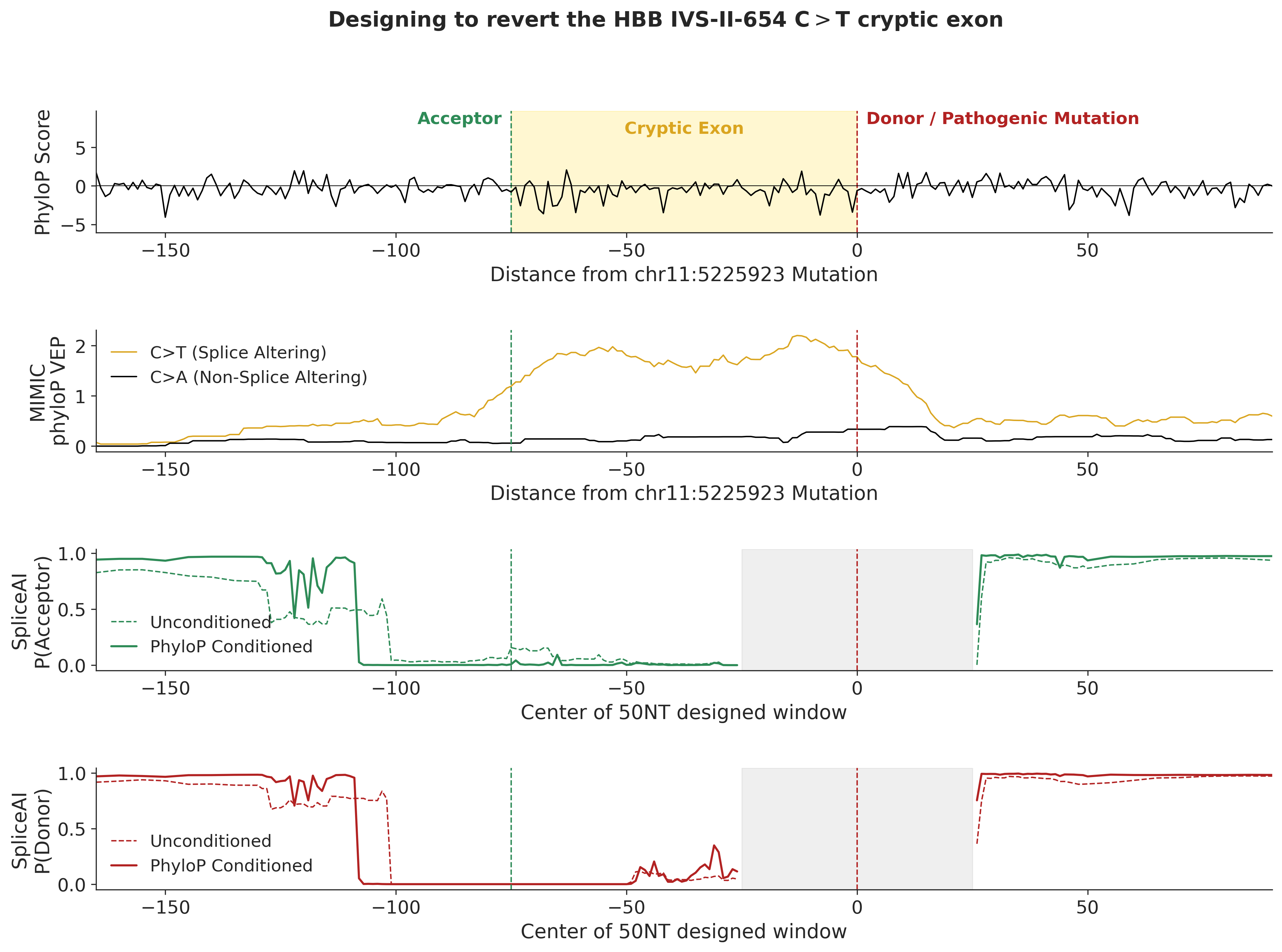}
    \caption{\textbf{Splicing design with a 50 nucleotide designable window.}}
    \label{fig:splice_design_50nt}
\end{figure}

Leaning on the strong RNA multimodal inpainting performance, we adopt a similar methodological approach for RNA design. We evaluate designed windows of both 30 nucleotides (\cref{fig:splicing}e) and 50 nucleotides (\cref{fig:splice_design_50nt}). For a given window of width $w$ centered at position $x$, we masked the interval $[x- \frac{w}{2}, x + \frac{w}{2}]$ from the full unspliced transcript sequence and provided this sequence as conditioning to MIMIC. In addition, we provided the full splice junctions track, with the wild type acceptor and donor specified, and (optionally) the full wild-type phyloP as conditioning. Designs were generated in a single model pass, where nucleotides were sampled from a softmax over the model predicted logits at each position (temperature $=1e-8$). We sample designs from windows centered every 5 nucleotides from $-300$ to $-130$, every 1 nucleotide from $-130$ to $+50$, and every 5 nucleotides from $+50$ to $+200$ from the pathogenic site. In order to avoid trivially reverting the pathogenic mutation, we avoid designing in a $\frac{w}{2}$-width window around the mutation location.

Designs are evaluated using SpliceAI \cite{Jaganathan2019SpliceAI} as an independent oracle, since MIMIC-predicted splicing would be circular with MIMIC designs. For each designed sequence, we query SpliceAI for the probability of an acceptor at the cryptic donor ($x=0$) and acceptor ($x=-75$) sites.

\subsection{Protein sequence design}
\label{app:prot_seq_design}

To explore the value of multimodal conditioning, we design amino acid sequences under five different conditioning cases: backbone only, surface features (MaSIF) for the 40\% of residues closest to the binding site (measured in 3D distance), all surface features (MaSIF), and the full backbone plus 40\%/100\% of MaSIF features. Backbone and surface features were computed using the downloaded PDB structures (PD-L1, 4ZQK chain A; hACE2, 6VW1 chain A).

We adopt an iterative design strategy wherein MIMIC is used for both generation and verification. In the generation step, a sequence is sampled from a softmax over logits from a single forward pass (temperature $=t$). We then prompt MIMIC with the designed sequence to predict the auxiliary modalities $\hat{M}$ (either backbone, surface features, or both). At each position $x$, we compute whether $\hat{M}_i$ matches the true conditioning modality $M_i$ with acceptance threshold $a$. For all positions such that $|\hat{M}_i -  M_i| > a$, these positions are marked for re-generation in the following iteration. We also randomly sample $r$ additional satisfied tokens for re-generation. Finally, with probability $p$ we resample positions in regions where the same amino acid is repeated $\geq 2$ times. This allows MIMIC flexibility to explore the design space without getting trapped in local maxima. Iteration proceeds for a maximum of 10 cycles, with the possibility to exit early if all positions are satisfied.

We treat this design problem as a hyperparameter search over temperature $t$, acceptance threshold $a$, and resampling rate $r$, and repeat resample probability $p$. We uniformly sample over \texttt{\{linear, cosine, exponential\}} annealing strategies from $t_{start} \in \{1, 1.5, 2, 3\}$ to $t_{end} \in \{0.001, \allowbreak 0.01, \allowbreak 0.05, \allowbreak 0.1,\allowbreak  0.5\}$, $a \in \{0.15, 0.2, 0.25, 0.3\}$, and \texttt{\{linear, cosine, exponential\}} annealing strategies from $r_{start} \in \{0.25, 0.3, 0.4, 0.5, 0.7, 0.9\}$ to $r_{end} \in \{0.02, 0.05, 0.08, 0.1\}$, and $p \in \{0.99, 1\}$. For each of the five conditioning cases described above, we randomly sample 200 hyperparameter sets and generate a designed sequence per set, then use ESMFold \cite{esm2} as a fast proxy of quality to filter to the top 20 designs for each case. 

Designs are evaluated along several axes. We first use AlphaFold2 to predict the fold for the generated sequence, filtering for high-confidence designs (average pLDDT > 85). To assess overall structural fidelity, we compute the TM-score of the AlphaFold2-predicted structure against the native structure. Alongside backbone topology, we quantify the conservation of surface properties by defining a MaSIF similarity score between the sequence-aligned predicted and native structures. To compute this score, we isolate jointly surface-exposed residues, retaining only positions mapped to at least one surface vertex in both structures. Across these shared surface residues, we calculate the Pearson correlation coefficient for each of the four biophysical features (shape index, electrostatic charge, hydrogen-bonding, and hydrophobicity). The final similarity score is the arithmetic mean of these four correlations, linearly rescaled to a range of 0 to 1. Consequently, a score of 1.0 indicates perfectly correlated surface chemistry, 0.0 indicates perfect anti-correlation, and 0.5 denotes no net correlation (this neutral value is also assigned by default if fewer than five jointly exposed residues are available for comparison). Finally, for these high-confidence designs, we evaluate their predicted binding using the AlphaFold3 web API and report the interface predicted template modeling score (iPTM) between chains.

\subsection{DNA preparation, RNA transcription, and purification}
The DNA template encoding each MIMIC construct was amplified by polymerase chain reaction (PCR) using a forward primer containing the T7 promoter sequence. Gene fragments for the full-length and 634~nt constructs were obtained from GenScript, and the template for the 274~nt constructs was assembled by PCR from three overlapping oligonucleotides (Supplementary Table~X). The transcription reaction consisted of 0.64~$\mu$M DNA template, 8~mM of each rNTP, 10\% polyethylene glycol 8000, 20--40~mM MgCl$_2$, $1\times$ transcription buffer (5~mM Tris, pH~8.0, 5~mM spermidine, 10~mM dithiothreitol [DTT]), and in-house expressed and purified T7 bacterial RNA polymerase. RNA was \textit{in vitro} transcribed at 37\,$^\circ$C for 3~h, then purified either by lithium chloride precipitation or by high-performance liquid chromatography (HPLC) on a DNAPac\texttrademark\ PA200 column (Thermo Scientific) using 12.5~mM Tris-HCl and 6~M urea and eluted with a gradient of 0--500~mM sodium perchlorate. Purified RNA was desalted using Sep-Pak\textregistered\ Plus Long C18 columns (Waters) or dialysis, lyophilized, and resuspended in nuclease-free water.

\subsection{2D structural probing}

\paragraph{Constructs used for \textit{in vitro} chemical probing.}
For direct comparison with secondary structure predictions from MIMIC, constructs consisting of the cryptic exon flanked by 75~nt on each side were used. An additional 25~nt corresponding to the adjacent transcript regions were included at both termini as primer-binding sites for PCR amplification prior to sequencing, as reactivity information cannot be recovered within primer-binding regions. The structures of the larger constructs used for \textit{in vitro} splicing assays were also chemically probed.
\begin{figure}
    \centering
    \includegraphics[width=1\linewidth]{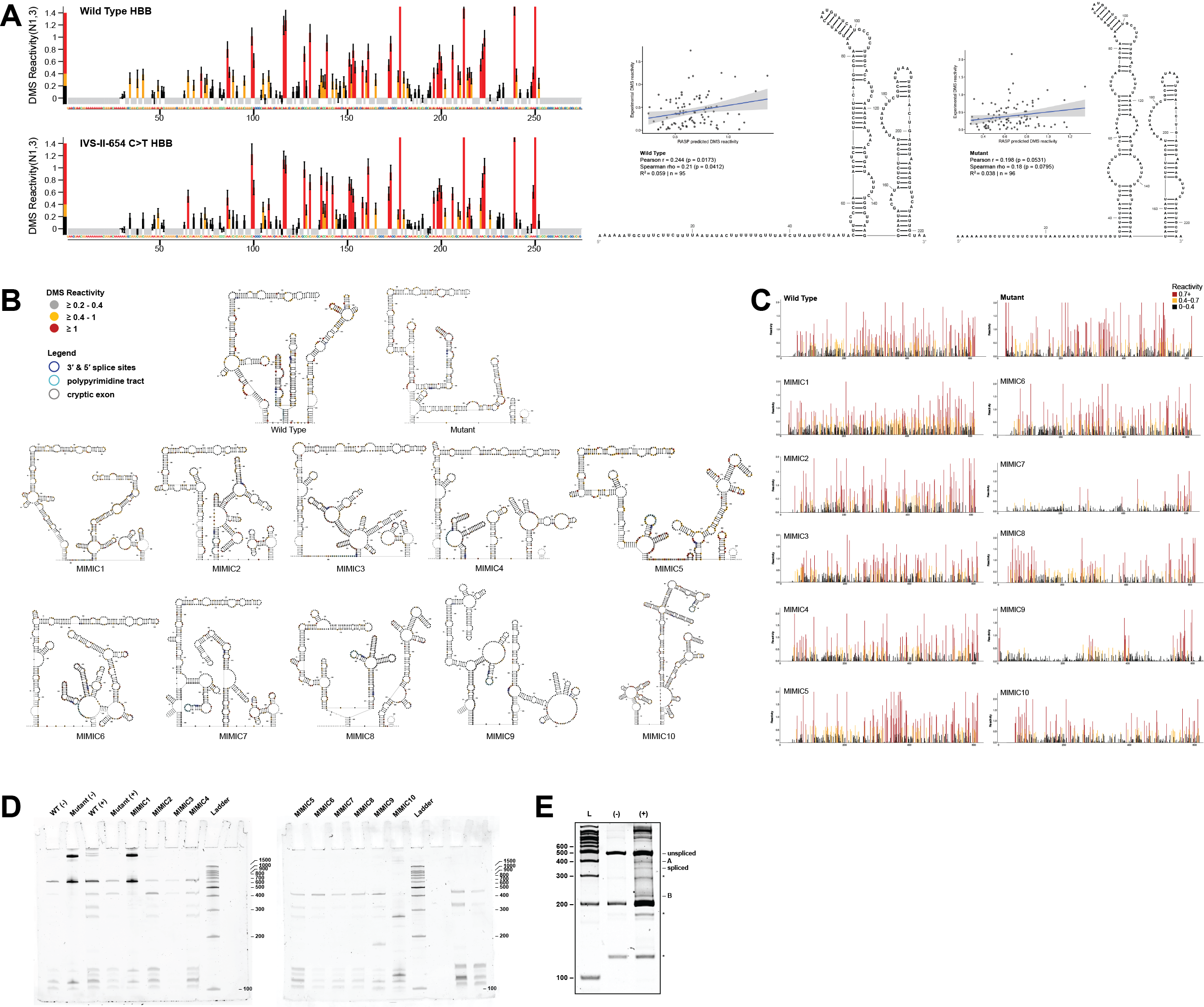}
    \caption{\footnotesize{\textbf{MIMIC predicts RNA secondary structures consistent with experimental measurements and guides construct design to mitigate mutation-induced splicing defects.} \textbf{(A)} DMS reactivity profiles for the HBB short constructs. Correlations between experimental and \texttt{rasp2} DMS reactivities for the WT and MUT short constructs are shown on the right. \textbf{(B)} Secondary structures of HBB and MIMIC rescues constructs used for in vitro splicing assays. \textbf{(C)} Reactivity profiles of HBB and MIMIC rescues constructs used for in vitro splicing assays. \textbf{(D)} Uncropped gels of in vitro splicing of HBB and MIMIC rescue constructs. \textbf{(E)} Gel of in vitro splicing of beta-globin showing presence of a band corresponding to the expected spliced product and additional splicing intermediates (A and B). (-) = no nuclear extract, (+) = nuclear extract, L = ladder, * = background amplicon products.}}
    \label{fig:placeholder}
\end{figure}
\paragraph{Chemical probing.}
Chemical probing was performed as described previously \cite{Zubradt2017, Smola2015, Arnold2025}. Briefly, \textit{in vitro} transcribed RNA was denatured at 95\,$^\circ$C for 5~min and immediately cooled on ice. RNA was refolded at 37\,$^\circ$C in a buffer containing 100~mM HEPES (pH~8.0), 10~mM MgCl$_2$, and 100~mM NaCl. For all chemical probing experiments, multiple probe concentrations were screened to ensure single-hit kinetics. For DMS-MaP experiments, RNA was treated with 25.3~mM dimethyl sulfate (DMS) at 37\,$^\circ$C for 15~min and quenched with 333~mM DTT. Modified RNA was purified using RNAClean XP beads (Beckman, A63987) and reverse transcribed using Induro Reverse Transcriptase (New England Biolabs [NEB], M0681L). The resulting cDNA was purified with RNAClean XP beads, amplified with Q5 Hot Start High-Fidelity DNA Polymerase (NEB, M0491S), and cleaned up using AMPure XP beads (Beckman, A63881). Libraries were sequenced on an Oxford Nanopore Technologies (ONT) Flongle Flow Cell (ONT, FLO-FLG114) using the Native Barcoding Kit (ONT, SQK-NBD114.24) \cite{Bose2024}. Sequencing reads for the short constructs were analyzed using ShapeMapper2 \cite{Busan2018}, whereas reads for the larger splicing constructs were analyzed using RNAFramework \cite{Incarnato2018}. RNAs were folded using \texttt{RNAfold} from the ViennaRNA package with DMS reactivity constraints to determine the minimum free-energy structure, and secondary structure diagrams were rendered using RNAcanvas \cite{Johnson2023}.

\subsection{Splice-site strength analysis}
Splice-site strength was evaluated using MaxEntPy, a Python implementation of the MaxEntScan maximum-entropy models for splice-site scoring (\url{https://github.com/kepbod/maxentpy}). For 5$'$ splice-site (5$'$SS) analysis, 9~nt sequences spanning positions $-3$ to $+6$ relative to the exon--intron junction (comprising the final 3~nt of the upstream exon and the first 6~nt of the intron) were evaluated. For 3$'$ splice-site (3$'$SS) analysis, 23~nt sequences spanning positions $-20$ to $+3$ relative to the intron--exon junction (comprising the final 20~nt of the intron and the first 3~nt of the downstream exon) were evaluated. Splice-site scores were computed using the \texttt{score5()} and \texttt{score3()} functions in MaxEntPy for 5$'$SS and 3$'$SS sequences, respectively, where higher MaxEnt scores indicate stronger concordance with canonical splice-site motifs.

\subsection{Nuclear extract preparation}
Nuclear extract (NE) was prepared from Jurkat Splicing Line 1 (JSL1) cells. Cells were harvested by centrifugation and washed twice with ice-cold PBS. Cell pellets were resuspended in five packed cell volumes (PCV) of Buffer~A (10~mM Tris-HCl, pH~7.5, 1.5~mM MgCl$_2$, 10~mM KCl, and 0.5~mM DTT) and incubated on ice. Following centrifugation, cells were resuspended in two PCV of Buffer~A and homogenized using a Dounce homogenizer. Nuclei were collected by centrifugation and resuspended in 0.7~PCV of Buffer~C (20~mM Tris-HCl, pH~7.5, 1.5~mM MgCl$_2$, 420~mM NaCl, 25\% glycerol, 0.2~mM EDTA, and 0.5~mM DTT), followed by Dounce homogenization. The suspension was gently rocked and centrifuged to remove insoluble material. The supernatant was dialyzed against BC100 buffer (20~mM Tris-HCl, pH~7.5, 100~mM KCl, 20\% glycerol, 0.2~mM EDTA, and 0.5~mM DTT) with one buffer change. Following dialysis, the NE was clarified by centrifugation, aliquoted, flash-frozen in liquid nitrogen, and stored at $-80\,^\circ$C until use.

\subsection{\textit{In vitro} splicing assays}
\textit{In vitro} splicing reactions were performed using JSL1 NE. RNA (1~$\mu$L) was mixed with 3.5~$\mu$L BC137 buffer in RNase-free tubes on ice. A splicing master mix containing creatine phosphate (CP; MilliporeSigma, 2380-5GM) and polyvinyl alcohol (PVA; Sigma-Aldrich, P8136-250G) was prepared separately on ice, with 13\% PVA added immediately prior to use; excess master mix was prepared to account for the high viscosity of PVA and ensure consistent pipetting across reactions. JSL1 NE was thawed immediately before use, and 4~$\mu$L was added to each reaction, followed by 4.5~$\mu$L of the splicing master mix. Reactions were incubated at 30\,$^\circ$C for 1.5--2~h. Globin was used as a positive control in every assay to confirm recombinant spliceosome activity (Figure S8e)

Following incubation, reactions were treated with recombinant PCR-grade Proteinase~K (Thermo Scientific, FEREO0492) at 37\,$^\circ$C for 15~min to digest proteins in the NE. RNA was recovered via phenol:chloroform:isoamyl alcohol (PCA; MilliporeSigma, 6805-100ML) extraction and ethanol precipitation: 200~$\mu$L PCA was added to each reaction, samples were vortexed and centrifuged for 5~min, and the upper aqueous phase was transferred to 500~$\mu$L ice-cold 100\% ethanol. Samples were incubated at $-80\,^\circ$C for 20~min and centrifuged at maximum speed for 15~min. RNA pellets were washed with ice-cold 70\% ethanol, centrifuged at maximum speed for an additional 5~min, air-dried after removing residual ethanol, and resuspended in nuclease-free water. Recovered RNA was reverse transcribed using Induro Reverse Transcriptase (NEB, M0681L), and the resulting cDNA was directly amplified by PCR using Q5 Hot Start DNA Polymerase (NEB, M0491S). Amplicons corresponding to unspliced and spliced products were resolved by polyacrylamide gel electrophoresis and visualized by SYBR Gold (Invitrogen, S11494) staining. All experiments were carried out in duplicate.

\end{document}